%% file: 00_main.tex
\documentclass{article}
\pdfoutput=1
\usepackage{arxiv}

\usepackage{float}
\usepackage{caption}
\usepackage{times}

\usepackage{multicol}
\usepackage{todonotes}
\usepackage{multirow}
\usepackage{booktabs}
\usepackage{makecell}
\usepackage{tabularx}
\usepackage{siunitx}
\usepackage{amsmath}
\usepackage{amssymb}
\usepackage{stfloats}
\usepackage{cuted}
\usepackage{placeins}
\usepackage{graphicx}
\usepackage[export]{adjustbox}
\usepackage{wrapfig}
\usepackage{paralist}
\usepackage{subcaption}
\usepackage{algorithmic}
\usepackage{algorithm}
\usepackage{tikz-network}
\usepackage{natbib}

\newcommand{\ie}{\textit{i}.\textit{e}. }
\newcommand{\eg}{\textit{e}.\textit{g}. }

\newcommand{\push}{\vspace{1em}}

\title{A Simple Approach to Continual Learning by Transferring Skill Parameters}

\author{
  K.R. Zentner$^*$,
  Ryan Julian\thanks{Equal contribution}$\;\:$,
  Ujjwal Puri,
  Yulun Zhang,
  and
  Gaurav S. Sukhatme
\thanks{All authors are with the Department of Computer Science,
         University of Southern California, Los Angeles, CA 90089.
         {\tt \footnotesize kzentner|rjulian|ujjwalpu|yulunzha|gaurav@usc.edu}. GS holds concurrent appointments as a Professor at USC and as an Amazon Scholar.}}

\begin{document}
\maketitle

%===============================================================================

\begin{abstract}
    In order to be effective general purpose machines in real world environments, robots not only will need to adapt their existing manipulation skills to new circumstances, they will need to acquire entirely new skills on-the-fly.
    A great promise of continual learning is to endow robots with this ability, by using their accumulated knowledge and experience from prior skills.
    We take a fresh look at this problem, by considering a setting in which the robot is limited to storing that knowledge and experience only in the form of learned skill policies.
    We show that storing skill policies, careful pre-training, and appropriately choosing when to transfer those skill policies is sufficient to build a continual learner in the context of robotic manipulation.
    We analyze which conditions are needed to transfer skills in the challenging Meta-World simulation benchmark.
    Using this analysis, we introduce a pair-wise metric relating skills that allows us to predict the effectiveness of skill transfer between tasks, and use it to reduce the problem of continual learning to curriculum selection.
    Given an appropriate curriculum, we show how to continually acquire robotic manipulation skills without forgetting, and using far fewer samples than needed to train them from scratch.
\end{abstract}

%===============================================================================
% The (one) big idea:
% Structured exploration is essential for efficient continual robot skill learning.
%
% To learn a new skill, it is more efficient to explore in the space of already-
% acquired skills than it is to explore in the action space.
%===============================================================================

\section{Introduction}
\label{sec:introduction}
\input{01_introduction.tex}
\section{Related Work}
\label{sec:related_work}
\input{02_related_work.tex}

% \section{Understanding the Role of Parameters and Behaviors in Skill Policy Reuse}
% \label{sec:understanding_reuse}
% \input{03_understanding_reuse.tex}

\section{Setting}
\label{sec:setting}
\input{03_setting.tex}

\section{Simple Continual Learning with Skill Transfer}
\label{sec:skill_transfer}
\input{04_skill_transfer.tex}

\section{Efficient Continual Learning with Skill Curriculum}
\label{sec:skill_curriculums}
\input{05_skill_curriculums.tex}

\FloatBarrier
\section{Conclusion}
\label{sec:conclusion}
\input{07_conclusion.tex}

\bibliographystyle{unsrtnat}
% \bibliography{references}  % .bib


\input{00_main.bbl}
\end{document}

%% file: 01_introduction.tex
%As we do not have complete information on all skills a robot might be asked to perform at design time, in the parlance of machine learning this setting is necessarily a continual learning problem.
%Specifically, we seek robots which can efficiently and continually learn new manipulation skills, while presumably making best use of skills they have already learned to do so.

% Differences from SL, which is why meta-learning has underwhelming results
% O(dozens) of tasks, not 1000s of examples
% Each task needs (expensive) exploration, rather than only collection and labeling

%
Reinforcement learning (RL) with rich function approximators---so-called ``deep'' reinforcement learning (DRL)---has been used in recent years to automate tasks which were previously-impossible with computers, such as beating humans in board and video games~\cite{mnih2013playing,mnih2015human} and navigating high-altitude baloons~\cite{bellemare2020autonomous}.
In the field of robotics, DRL has shown the promise by allowing robots to automatically learn sophisticated manipulation behaviors end-to-end from high-dimensional multi-modal sensor streams~\cite{levine2016end}, and to quickly adapt these behaviors to new environments and circumstances~\cite{julian2020never}.
What remains to be seen is whether DRL can bridge the significant gap from efficiently adapting existing skills to efficiently acquiring entirely new skills.
If such a capability could be applied repeatedly throughout the life of a robot (\ie continual learning), we stand the chance of unlocking new possibilities for physical automation with general purpose robots, much as general purpose computers unlocked theretofore unforeseen possibilities for information automation half a century ago.

While not the only relevant formulation, we believe that episodic, continual, multi-task reinforcement learning is a worthy problem setting, because it describes this skill acquisition capability we seek.
In  this setting, we ask the robot to acquire new manipulation skills repeatedly, using time-delineated experiences of attempts at those skills (episodes), and some durable store of previously-acquired knowledge.
The possibilities for the form of this store seem endless, but are actually bound to only two possibilities by construction: an RL system consumes raw data in the form of \textit{experiences}, and outputs processed data in the form of \textit{parameters}, which either directly specify a policy function, or condition a policy decision rule by specifying one or more other functions (\eg a value function, Q-function, transition model, etc.).
% \footnote{Neglecting the degenerate case of a policy computed just-in-time from raw data.} 
So a continual reinforcement learning robot can store one or both of (1) experience data or (2) parameters.

Among these two options, there are many reasons to prefer parameters over data.
Keeping a comprehensive dataset of all prior experience in local storage quickly becomes intractable for a single robot.
Even if stored remotely in a ``cloud'' and retrieved, no server could quickly locate and retrieve a subset of that data relevant to a new task without first processing it into parameters itself.
Parameters not only allow a single robot to store all of its skills locally, they are also more practical to share and disseminate than datasets (precisely because they are already processed).
Consider the success of pre-trained models in language and computer vision: these are computed at great expense by institutions with immense datasets, storage, and compute resources.
These institutions share them for the benefit of the entire community, who can then quickly re-use them for myriad applications.
The parameters for state-of-the-art language and computer vision models fit on a cell phone in the palm of your hand, but require warehouse-sized machines to compute.
While the best continual learning robot will likely store a mix of both data and parameters, we believe parameters deserve an especially-enthusiastic study.
For the sake of simplicity, in this work we shall focus on them in isolation.

% If we imagine a continual learning robot which can store only skill policies but no experiences, how can it use them for acquiring new skills?
% Skill policies have two uses: their \textit{parameters} directly and---uniquely to reinforcement learning---the \textit{behaviors} they indirectly encode.

In this work, will show that under this skill storage-only assumption, efficient skill acquisition can be performed if the appropriate skills to transfer to each new task are known.
We formalize our continual learning setting in Section~\ref{sec:setting}, and describe how it maps onto the simulated robotic manipulation benchmark Meta-World.
% We then show how minor modifications to an ``off-the-shelf'' deep RL algorithm can allow for reliable, if inefficient, continual learning.
% Then, in Section~\ref{sec:skill_curriculums}, we show explore the impact of selecting the appropriate skills to transfer.
% requires using both skills policys' parameters \textit{and} the behaviors they encode.
% In Section~\ref{sec:understanding_reuse}, we first use experiments with imitation learning and fine-tuning to gain insight into how we can re-use both skill policy parameters and their behaviors to acquire new skills rapidly, by exploiting shared structure between tasks.
%including precisely defining how to transfer parameters.
%, and describe a simple continual learning method which uses an expanding library of skill policies as its knowledge store, and on-policy fine-tuning with Proximal Policy Optimization (PPO)~\cite{schulman2017proximal} for new skill acquisition.
%We call this simple approach ``Skill Builder'' because it acquires new skills by building on old ones, such that learning a new skill can never cause the robot to forget an old one.
%Using experiments with the challenging Meta-World benchmark, we show that Skill Builder can achieve continual learning for robotic manipulation skills.
In Section~\ref{sec:skill_curriculums} we investigate the precise conditions required to both transfer old skills and learn new ones using on-policy fine-tuning with Proximal Policy Optimization (PPO)~\cite{schulman2017proximal}.
We introduce a measure across ordered pairs of tasks that describes how efficiently on-policy transfer can learn a new task using skills from a single prior task.
We then show how this measure allows constructing a curriculum predicted to be at least as efficient as learning each task without transfer, and we empirically verify that these predictions hold on the Meta-World benchmark.

%% file: 02_related_work.tex
\paragraph{Reinforcement learning for robotics}
Reinforcement learning has been studied for decades as an approach for learning robotic capabilities~\cite{kober2013reinforcement,mahadevan1992automatic,lin1992reinforcement,smart2002effective}.
In addition to manipulation skills~\cite{levine2018robotarmy,kalashnikov2018scalable,pinto2016supersizing,gullapalli1994acquiring,ghadirzadeh2017deep,zeng2018learning}, RL has been used for learning locomotion~\cite{kohl2004policy,stone2004machine,xie2019iterative,haarnoja2019learning}, navigation~\cite{beom1995sensor,zhu2017target}, motion planning~\cite{singh1994robust,everett2018motion}, autonomous helicopter flight~\cite{bagnell2001autonomous,abbeel2007application,ng2003autonomous}, and multi-robot coordination~\cite{mataric1997reinforcement,yang2004multiagent,long2018towards}.
The recent resurgence of interest in neural networks for use in supervised learning domains such as computer vision and natural language processing, (\ie ``deep learning'' (DL))~\cite{bengio2017deep}, corresponded with a resurgence of interest in neural networks for reinforcement learning (\ie ``deep reinforcement learning'' (DRL))~\cite{franccois2018introduction,mnih2013playing}.
With it came a wave of new research on using RL for learning in robotics and continuous control~\cite{mnih2015human,lillicrap2015continuous}, though the fields of neural networks, reinforcement learning, and robotics have overlapped continuously since each of their inceptions~\cite{kober2013reinforcement,hadsell2009learning}.

\paragraph{Transfer, continual, and lifelong learning for robotics}
Transfer learning is a heavily-studied problem outside the robotics domain~\cite{donahue2014decaf,ulmfit,devlin2018bert,dai2007boosting,raina2007self}.
Many approaches have been proposed for rapid transfer of robot skill policies to new domains, including residual policy learning~\cite{silver2018residual}, simultaneously learning across multiple goals and tasks~\cite{ruder2017overview,rusu2016progressive}, methods which use model-based RL~\cite{finn2017deep,yen2019experience,nagabandi2019deep,chatzilygeroudis2018using,ha2018recurrent,dasari2019robonet,chatzilygeroudis2018reset,cully2015robots,kaushik2020adaptive,merel2019reusable,rastogi2018sample, jeong2019modelling}, and goal-conditioned RL~\cite{agrawal2016learning,nair2018visual,Pathak_2018,pong2019skew,yu2019unsupervised}.
All of these share data and representations across multiple goals and objects, but not skills per se.
Similarly, work in robotic meta-learning focuses on learning representations which can be quickly adapted to new dynamics~\cite{nagabandi2018learning,alet2018modular,nagabandi2018deep} and objects~\cite{finn2017one,james2018task,yu2018one,bonardi2019learning}, but has thus far been less successful for skill-skill transfer~\cite{yu2019meta}.
Pre-training methods are particularly popular, including pre-training with supervised learning~\cite{deng2009imagenet,levine2016end,finn2016deep,gupta2018robot,pinto2016supersizing}, experience in simulation~\cite{sadeghi2017cadrl,tobin2017domain,SadeghiTJL18,sim2real,openai2019solving,Rusu2016SimtoRealRL,peng2018sim,higuera2017adapting,hamalainen2019affordance}, auxiliary losses~\cite{Riedmiller2018LearningBP,mirowski2016learning,sax2019mid}, and other methods~\cite{Sermanet2017Rewards,hazara2019transferring}.
While successful, these methods are often designed for domain transfer rather than skill-skill transfer, require significant engineering by hand to anticipate specific domain shifts, and are designed for single-step rather than continual transfer.
Similar to~\citeauthor{julian2020never} and~\citeauthor{nair2020accelerating}, our work uses the very simple approach of on-line fine-tuning to achieve rapid adaptation.

Lifelong and continual learning have long been recognized as an important capability for autonomous robotics~\cite{thrun1995lifelong}.
Like~\citeauthor{taylor2007transfer}, our approach to continual learning relies on rapidly adapting policies for an already-acquired skill into a policy for a new skill.
Much like~\citeauthor{cao2021transfer},~\citeauthor{bodnar2020geometric}, and~\citeauthor{kumar2020one}, this work uses experiments to analyze different transfer techniques from a geometric perspective on the skill-skill adaptation problem.
As in prior work~\cite{luna2020information,yen2020learning}, this study observes that the selection of pre-training tasks is essential for preparing RL agents for rapid adaptation.
Our work uses experiments to formulate a decision rule for how to pre-train our skills.
A comprehensive overview of literature in continual reinforcement learning beyond robotics is beyond the scope of this work, but please see~\citeauthor{khetarpal2020towards} for an excellent survey.

\paragraph{Reusable skill libraries for efficient learning and transfer}
Learning reusable skill libraries is a classic approach~\cite{gullapalli1994acquiring} for efficient acquisition and transfer of robot motion policies.
Prior to the popularity of DRL-based methods, Associative Skill Memories~\cite{pastor2012asm} and Probabilistic Movement Primitives~\cite{rueckert2015movprim,zhou2020incremental} were proposed for acquiring a set of reusable skills for robotic manipulation.
In addition to manipulation~\cite{tanneberg2021skid,yang2020multi,ichter2020broadly,wulfmeier2020data,vezzani2020not,camacho2020disentangled,li2021solving,lu2021learning,kroemer2015towards}, DRL-based skill decomposition methods are particularly popular today for learning and adaptation in locomotion and whole-body humanoid control~\cite{peng2019mcp,hasenclever2020complementary,merel2020catch,li2020learning,tirumala2020behavior}.
Our work argues that once decomposed, these skill libraries are useful for rapid adaptation, and ultimately continual learning for manipulation with real robots.
~\citeauthor{hausman2018learning} proposed learning reusable libraries of robotic manipulation skills in simulation using RL and learned latent spaces, and~\citeauthor{julian2018scaling} showed these skill latent spaces could be used for efficient simulation-to-real transfer and rapid hierarchical task acquisition with real robots.
As we also study in this work, learning reusable skill libraries requires exploring how new skills are related to old ones.
As other works have pointed out~\cite{benureau2016behavioral,singh2020parrot,biza2021action,singh2020cog,allshire2021laser}, we believe this can be achieved efficiently by re-using policies, representations, and data from already-acquired skills.

\paragraph{Continual robot learning with skill libraries and curriculums}
Like ours, recent works have begun to use DRL with skill libraries for continual robot learning.
They have explored maintaining a skill library in form of factorized policy model classes~\cite{mendez2020lifelong}, learned latent spaces~\cite{lu2020reset,koenig2017robot,hazara2019active}, options~\cite{hawasly2013lifelong}, policy models which partition the state space~\cite{xiong2021state}, movement primitives~\cite{maeda2017active}, or as per-skill or all-skill datasets~\cite{traore2019discorl,lu2020reset,hazara2019active}.
As in our work and others~\cite{traore2019discorl,stulp2012adaptive}, \citeauthor{fernandez2006probabilistic} proposed directly storing and re-using policies for continual learning in the context of robot soccer.
Once acquired, these works propose various methods for reusing these skills, such as via online model-based planning~\cite{lu2020reset}, via sequencing, mixture, selection, or generation with online inference~\cite{xiong2021state,stulp2012adaptive,hazara2019active,maeda2017active}, as a high-level action space for hierarchical RL~\cite{hawasly2013lifelong}, and (as in our work) keeping a specific policy network for each new skill~\cite{mendez2020lifelong,traore2019discorl,koenig2017robot}.
Intertwined with how to maintain such skill libraries is the question of how to update them throughout the life of the robot.
Recent works have proposed using on-policy RL algorithms to directly update skills~\cite{mendez2020lifelong,xiong2021state,stulp2012adaptive,koenig2017robot,maeda2017active}, using a continually-growing skill data buffer to update skill networks~\cite{lu2020reset,hazara2019active}, and repeatedly distilling the policy library~\cite{hawasly2013lifelong,traore2019discorl}.

We believe that continual learning for manipulation is achievable by using modular skill libraries and repeated efficient adaptation to new tasks.~\citeauthor{alet2018modular},~\citeauthor{sharma2020learning}, and~\citeauthor{raziei2020adaptable} have all recently proposed rapid adaptation methods for manipulation which make use of modular skill learning and re-use.
This work seeks to extend some of those ideas, in simplified form, to the continual learning setting.
% In addition to our analysis of skill policy transfer and skill curriculums, we propose policy class and pre-training procedure for skill re-use which is inspired by~\citeauthor{peng2019mcp} and~\citeauthor{tseng2021toward}, but for continual reinforcement learning rather than imitation learning or meta-RL.

See~\citeauthor{narvekar2020curriculum} for a survey of curriculum learning in reinforcement learning.
Like~\citeauthor{fabisch2015accounting}, our work observes that continual skill learning is an active learning problem, and that measuring task novelty is an important capability for efficient active skill learning.
Like and~\citeauthor{foglino2019curriculum}, we highlight the importance of skill curriculum, and propose a method for computing the optimal skill curriculum given an oracle for relative skill novelty, and show that these curriculums indeed make continual learning more efficient.

%% file: 03_setting.tex
% In this section, we establish our formal continual learning setting and how it maps onto the Meta-World benchmark~\cite{yu2019meta}.
% We then describe an incredibly simple approach to continual learning without forgetting, by performing on-policy finetuning of policies from prior tasks using Proximal Policy Optimization (PPO)~\cite{schulman2017proximal}.

%fine-tuning existing policies using
%, and introduce Skill Builder, our simple approach to continual learning for manipulation.
%We use experiments with the challenging Meta-World manipulation benchmark~\cite{yu2019meta} to establish that Skill Builder can achieve continual learning without forgetting.
%Later in Section~\ref{sec:skill_curriculums}, we will show how skill curriculums can make continual learning with Skill Builder more efficient than learning skills from scratch.

% \subsection{Problem Setting}
We formalize our continual learning problem as iterated transfer learning for multi-task reinforcement learning (MTRL) on a possibly-unbounded discrete space of tasks $\mathcal{T}$.
As we are interested in learning robot manipulation policies, we presume all tasks in $\mathcal{T}$ share a single continuous state space $\mathcal{S}$ and continuous action space $\mathcal{A}$, and the MTRL problem is defined by the tuple $\left(\mathcal{T}, \mathcal{S}, \mathcal{A}\right)$
Each task $\tau \in \mathcal{T}$ is an infinite-horizon Markov decision process (MDP) defined by the tuple $\tau = \left(\mathcal{S}, \mathcal{A}, p_\tau(s,a,s'), r_\tau(s,a,s')\right)$.
As tasks are differentiated only by their reward functions $r_\tau$ and state transition dynamics $p_\tau$, we may abbreviate this definition to simply $\tau = (r_\tau, p_\tau)$.

% \todo{KR: We need to talk about pre-training here.}
Importantly, we do not presume that the robot ever has access to all tasks in $\mathcal{T}$ at once, or even a representative sample thereof, and can only access one task at a time.
We shall refer to time between task transitions an ``epoch'' and count them from 0, but in general two different epochs can be assigned the same task (\ie tasks may reappear).
When solving a task $\tau$ (hereafter, the ``target task''), the robot only has access to skill policies acquired while solving prior tasks $\mathcal{M}$ (the ``skill library'').
When only a single prior task is used to solve a new task, we will refer to that task as the ``prior task.''
Extending our problem to include these assumptions, we can say that a single epoch of this continual multi-task learning problem is defined by an infinite-horizon MDP $\left(\mathcal{S}, \mathcal{A}, \mathcal{M}_i, p_{\tau_i}, r_{\tau_i}\right)$, where $i$ is the epoch number and $\mathcal{M}_0$ is the (possibly-empty) set of manipulation skills with which the robot is initialized.

In this work, we will assume that the robot can choose which task $\tau$ to learn in each epoch, and also when to stop learning that task and begin a new epoch.
In Section~\ref{sec:skill_curriculums} we will discuss at length the implications of such decisions.
% , and in Section~\ref{sec:towards_skill_policy}, we will discuss some implications of how $\mathcal{M}_0$ is chosen.

%% file: 04_skill_transfer.tex
% \subsection{Simple Continual Learning with Skill-Skill Transfer}
%\begin{figure}
    %\centering
    %\includegraphics[height=2in, width=\textwidth]{figures/skillbuilder_freehand.png}
    %\caption{Overview diagram of for Skill Builder.
    %Starting with a library of pre-trained skills, the robot acquires a new skill by selecting an existing skill from its library and fine-tuning it with PPO~\cite{schulman2017ppo} in the new training environment.
  %Once acquired, the new skill is added to the library, and the robot repeats the process.}
    %\label{fig:skill_builder_diagram}
%\end{figure}

%\todo{KR: Is this actually the right order? We're no longer focused on PPO later in the paper.}
%Having convinced ourselves in Section~\ref{sec:understanding_reuse} that efficient skill reuse is possible with RL fine-tuning, we introduce a very simple framework we call Skill Builder (Figure~\ref{fig:skill_builder_diagram}).
%A Skill Builder starts with a pre-trained set of base skills $\mathcal{M}_0$.
%On each epoch $i$, it chooses a target task $\tau$ and a base skill policy $\pi_{base} \in \mathcal{M}_i$, and uses an RL algorithm $\mathcal{F}$ to fine-tune a clone of $\pi_{base}$ to solve $\tau$, which returns a new policy $\pi_{target}$ and the success rate $\rho$.
%It then adds $\pi_{target}$ to the skill library, and continues (Algorithm~\ref{alg:skill_builder}).
%\todo{KR: We need to work early-stopping into this part of the paper.}
Now that we have defined our setting in detail, we can describe our proposed continual learning procedure in abstract.
We begin with a (potentially empty) set of pre-trained skills $\mathcal{M}_0$.
Then, in each epoch $i$, we choose a target task $\tau$ and base skill policy $\pi_{base} \in \mathcal{M}_i$.
We then run an algorithm $\mathcal{F}$ to train a clone of $\pi_{base}$ to solve $\tau$.
This may either accept $\pi_{base}$ and return a new policy $\pi_{target}$ or reject the selected $\pi_{base}$, in which case a new $\tau$ and $\pi_{base}$ is chosen.

\hspace*{-\parindent}%
\begin{minipage}{\columnwidth}
\vspace{0.5em}
\begin{algorithm}[H]
    \begin{algorithmic}[1]
      \STATE {\bfseries Input:} Initial skill library $\mathcal{M}_0$, target task space $\mathcal{T}$, RL algorithm $\mathcal{F} \to (\pi, \rho)$, target task rule $\mathtt{ChooseTargetTask}$, base skill rule $\mathtt{ChooseBaseSkill}$
      \STATE $i \gets 1$
      \WHILE{not done}
        \STATE $\tau \gets \mathtt{ChooseTargetTask}(\mathcal{T}, \mathcal{M}_{i-1})$
        \WHILE{$\pi_{target}$ not solved}
          \STATE $\pi_{base} \gets \mathtt{ChooseBaseSkill}(\mathcal{T}, \mathcal{M}_{i-1})$
          \STATE $\pi_{target}, \cdot \gets \mathcal{F}(\tau, \mathtt{clone}(\pi_{base}))$
        \ENDWHILE
        \STATE $\mathcal{M}_{i} \gets \{\pi_{target}\}\cup \mathcal{M}_{i-1}$
        \STATE $i \gets i + 1$
      \ENDWHILE
      \STATE {\bfseries Output:} Skill library $\mathcal{M}_i$
    \end{algorithmic}
    \caption{Proposed Continual Learning Framework}
   \label{alg:skill_builder}
\end{algorithm}
\vspace{0.5em}
\end{minipage}

In this work, we primarily use PPO for the RL algorithm $\mathcal{F}$, but in general $\mathcal{F}$ may be any parametric RL algorithm, including an off-policy algorithm.
However, we find that we need to augment PPO in a few minor ways in order for it to perform adequately.
These augmentations are general enough that they could be applied to most on-policy DRL algorithms.

\paragraph{``Warm-Up'' Procedure for Value Function Transfer}
In order to tune a skill on a task using on-policy reinforcement learning, we also need a value function $V_{\tau,pi_{\tau'}}(s)$ which estimates the expected return of that skill policy on the task $\tau$.
Although PPO learns a value function as it trains the policy, we found that using a value function not fitted to the current task destroyed the skill policy's parameters before they could be transferred by PPO.
Copying value functions of the same skill on prior tasks is particularly ineffective, since those value functions are very likely to over-estimate initial performance, and are thus not admissible.
To avoid this issue, before applying any gradient updates to the policy, we sample a batch of the skill's behavior on the new task, and train a new value function to convergence on the Monte Carlo return estimates from those samples.
We found that this ``value function warm-up'' procedure was sufficient to produce an accurate value function of any skill on the new task, and was necessary to perform transfer effectively with PPO.

\paragraph{Rejecting Bad Transfers}
With Value Function Warm-Up, it is possible to transfer a skill policy $\pi_{base}$ to a new task $\tau$.
However, this process is never completely reliable.
If $\pi_{base}$ is particularly unsuitable for solving $\tau$, then PPO may be completely unable to transfer $\pi_{base}$ to solve $\tau$.
Because we desire a \textit{continual} learner, we must be able to continue learning without spending too much time on bad transfers.
To this end, we make use of a rejecting rule that can stop training at any point in time, and request a new $\pi_{base}$.
In this work, we use a rejection rule that compares the current average reward of $\pi_{base}$ on $\tau$ to the average reward of training a policy from-scratch to solve $\tau$ using 1.2 million ($10\%$ of total training time) fewer timesteps than have been used so far in the transfer process.
If the transferred policy $\pi_{base}$ ever ``falls too far behind'' the from-scratch policy, then we reject $\pi_{base}$ and select a new $\pi_{base}$ (possibly $\pi_{random}$).

\subsection{Using Meta-World for Continual Learning}

\begin{figure}[H]
    \centering
    \includegraphics[width=\columnwidth]{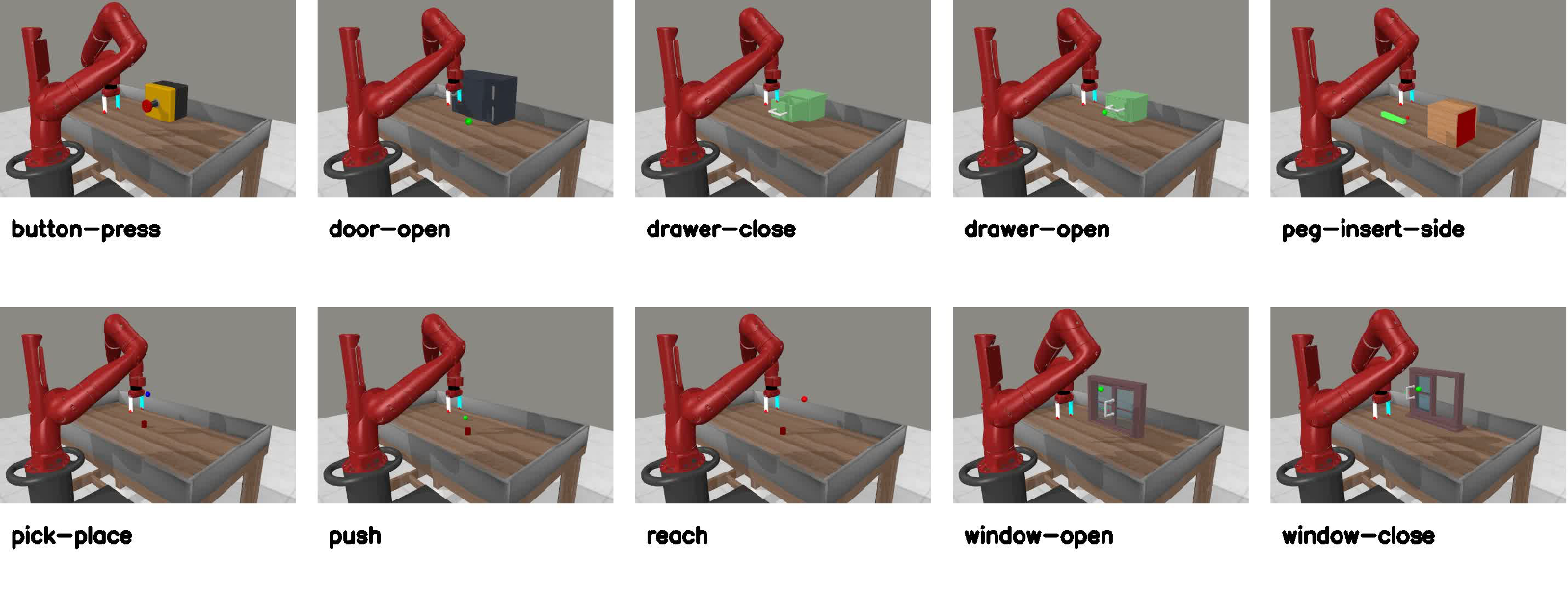}
    \caption{Images of Meta-World MT10 environments we use for experiments.
    }
    \label{fig:metaworld_envs}
\end{figure}

In order to perform useful continual learning experiments for robotic manipulation, we need a benchmark that allows us to learn multiple distinct robotic manipulation tasks.
In this work we use environments from the Meta-World MT10 benchmarks.
In particular, we consider each distinct environment in Meta-World as a seperate task (e.g. \texttt{pick-place}, \texttt{reach}, \texttt{push}, etc.).
We use the fully observable variants of those environments, as found in the MT10 benchmark.
This ensures that each ``task'' in our experiments internally contains parametric variation that skill policies must encode how to handle.

\subsection{Learning Continually using Random Skill Transfer}
With these augmentations, we can learn continually (if inefficiently), by transferring skills from random prior tasks.
The performance of this method is equivalent to a ``random curriculum'' as presented in the next section.
See Figure~\ref{fig:curriculum_frontier} for details on the performance of Random Skill Transfer.

%% file: 05_skill_curriculums.tex
Now that we have show that Random Skill Transfer can produce a continual learner that can learn by only transferring skill policies from one task to another, we would like to make a continual learner that is more efficient than from-scratch learning.
We do this by constructing a ``curriculum'', which will be used by the function $\mathtt{ChooseBaseSkill}$ to choose a base skill for each target task.
In the next Section, we will introduce a skill policy model for online curriculum selection, and share some promising results towards using it for continual learning.

We first develop a notion of skill-skill transfer cost, by counting the number of samples needed to acquire a target skill starting from a given base skill.

We then show that given this metric, we can reduce efficient continual learning to solving a Directed Minimum Spanning Tree (DMST) problem.
We show the effectiveness of our skill curriculum selection algorithm by using it to continually learn all skills in the Meta-World MT10 benchmark using a fraction of the total samples needed to learn each skill from scratch.

\subsection{Measuring Skill-Skill Transfer Cost}
We define efficiency in continual learning as acquiring a skill policies for a given set of tasks while consuming the lowest number of environment samples possible.

Since we are reducing the continual multi-task learning problem to one of repeated skill-skill transfer, it follows that efficient continual learning is equivalent to minimizing the sum of the environment steps used for each adaptation step.
Without any prior on the relationship between two tasks, estimating such a quantity is difficult~\cite{sinapov2015learning}.
This is compounded by the fact that skill-skill transitions are not independent: the robot only has access to skill policies it acquired during adaptation to tasks it has already seen, so the lowest-cost skill-skill transition for any given epoch depends on the skills which were acquired in previous epochs.

%Inferring the skill-skill transfer cost for a pair of tasks offline is itself a rich research problem we leave for another day.
To make progress on our central question, we make two simplifying assumptions: (1) before continual learning begins, we can access the task space to build a cost metric for skill-skill transfer and (2) we assume that skill-skill transfer costs are conditionally-independent (\ie as long as the robot has a skill policy for a manipulation task, the skill-skill transfer cost is independent of the skill transfer sequence it used to acquire that policy).
Our experiments below with Meta-World MT10 will validate the conditional independence assumption.
% \todo{KR: reword references to Section~\ref{sec:towards_skill_policy}}
% In Section~\ref{sec:towards_skill_policy} we introduce a skill policy class which shows promising results towards inferring the skill-skill transfer cost online.

% However, searching across all of those sequences to find an optimal one is not practical.
% To make finding an optimal curriculum feasible, we assume that there is a fixed cost to transfer skills learned on one task to another task.

To build our offline skill-skill cost metric, before continual learning begins, we train a skill policy from scratch for each task $\tau \in \mathcal{T}$.
We then use these as base skill policies, and retrain a copy of each to solve each other target task.
We considered several ways we could define the cost of transferring a skill to a new task.
Our initial experiments used the inverse of the average success rate throughout a fixed training interval as the cost.
This metric is convenient because it is always well defined, even if the transferred skill fails to learn the new task.
However, because transferred skills often exhibits a sudden improvement in performance after a fixed number of environment timesteps, this metric is highly sensitive to the size of the training interval used to compute the performance ratio.
It also does not accurately represent the resources needed in the learning process (namely, the number of environment steps needed to train).
Because most skills that succeed during transfer eventually reach a $>90\%$ success rate, we chose to use that threshold as a performance criteria instead, and use the number of timesteps required to reach it as the skill-skill transfer cost $C$ (See Equation~\ref{eq:cost_metric}).
This is similar to the ``jumpstart'' measure used for skill curriculum inference by~\citeauthor{sinapov2015learning}.

Note that in \cite{yu2019meta}, several tasks which we include in our experiments cannot be reliably solved to this success rate.
However, we find that our restarting rule allows us to learn all skills we include in our experiments to the required success rate.

\begin{figure}[H]
        \centering
        \includegraphics[width=\columnwidth]{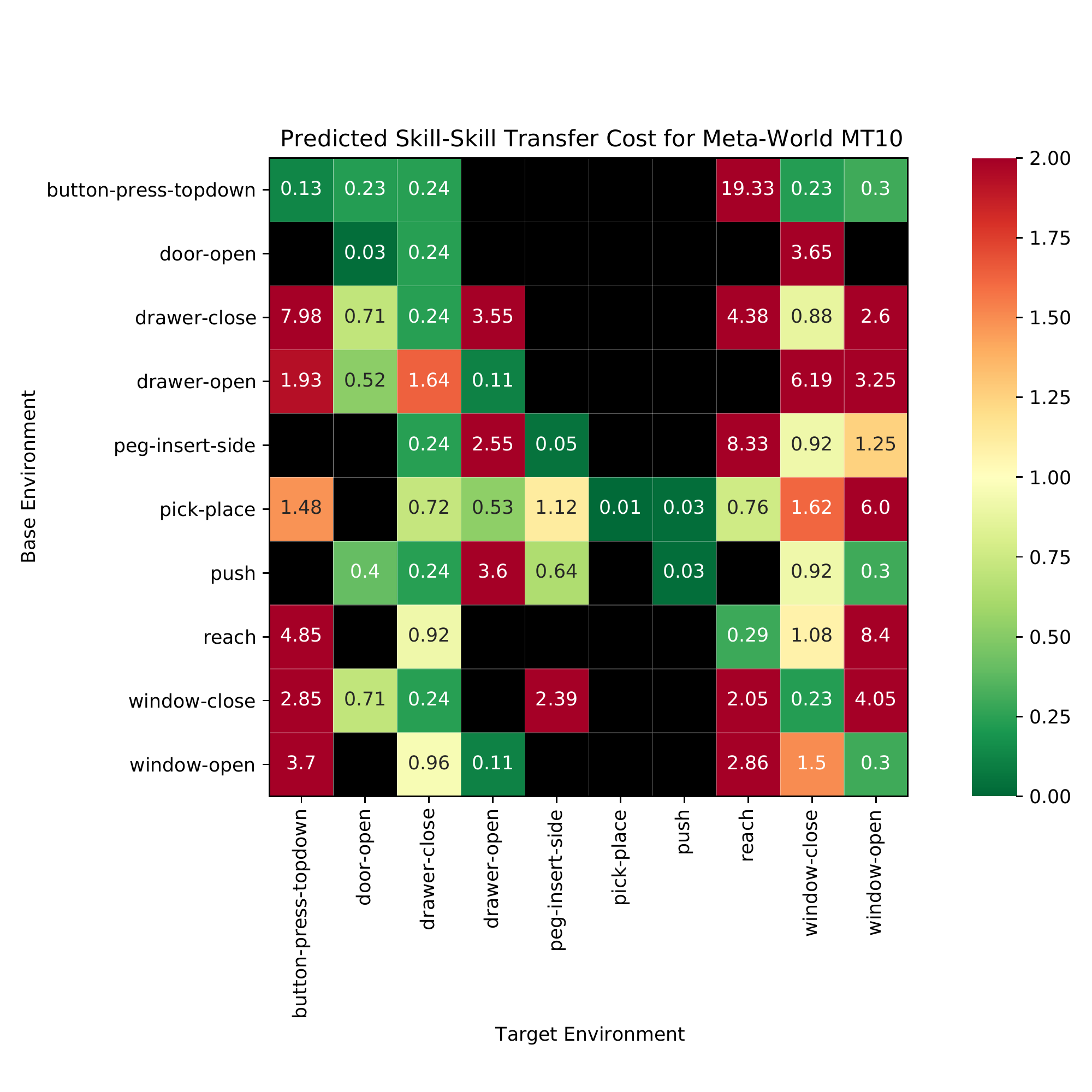}
        \caption{
        This figure shows $A_{base \to target}$, as defined in Equation~\ref{eq:cost_metric},
         the ratio of time steps required to learn a task in MT10 by skill transfer to learning the same task from scratch, using each other possible skill policy as a base skill.
        Note that we always run at least a single training iteration, and use at most 12 million timesteps.
        This prevents the matrix from containing 0 along the diagonal.
        Black cells in the matrix correspond to transfers that did not reach a $90\%$ success rate.
        }
        \label{fig:mt10_matrix}
\end{figure}

\begin{align}
\begin{split}
    C_{base \to target} &= \textsc{EnvironmentSteps}_{base \to target} \\
    C_{scratch \to target} &= \textsc{EnvironmentSteps}_{scratch \to target} \\
    A_{base \to target} &= \frac{C_{base \to target}} {C_{scratch \to target}}
\end{split}
\label{eq:cost_metric}
\end{align}

\subsection{Curriculum Selection Algorithm}
 Given a skill-skill transfer cost for each task $\tau \in \mathcal{T}$, we can generate a curriculum which minimizes the transfer cost of learning each $\tau \in \mathcal{T}$.
% Iterated fine-tuning requires a way of selecting the prior skills to be transferred when a new task is encountered.
We refer to the choice of skill to transfer on each task as ``the curriculum.''
Our curriculum selection algorithm is based on the observation that we can interpret the skill-skill transfer cost matrix in Figure~\ref{fig:mt10_matrix} (which shows $A_{base \to target}$) as the weighted adjacency matrix of a densely-connected directed graph, with our skill-skill transfer cost metric $C_{base \to target}$ as the directed edge weights.
Under this interpretation, we can extract the lowest cost of visiting all tasks by solving for the Directed  Minimum Spanning Tree (DMST) using Kruskal's Algorithm~\cite{kruskal1956shortest}.
The resulting tree will have a total edge weight equal to the minimal number of environment steps required to learn all tasks, as predicted by the skill-skill transfer cost metric.

To complete this DMST formulation, we need to take into account the possibility that it is most efficient to train a skill policy from scratch, rather than starting learning with one which already exists in the skill library $\mathcal{M}$.
To achieve this, we add a ``scratch'' vertex to our graph representation, with out-directed edges from the scratch vertex to each task with edge weight equal to the number of training steps needed to learn that task from scratch.
% $-1$, representing no advantage (\ie a ratio of 1) compared to training from scratch.
This would be represented in Figure~\ref{fig:mt10_matrix} as an addition row, whose values are all $1.0$.

\begin{figure}[h]
    \scalebox{0.9}{
      \begin{tikzpicture}
        \clip (0,4.5) rectangle (9,13);
        \Vertex[x=4.5,y=12,size=1,label=scratch,shape=rectangle]{scratch}
        \Vertex[x=1.5,y=10.5,size=1,label=pick-place,shape=ellipse,shape=ellipse,style={minimum width=70}]{pick-place}
        \Vertex[x=1.5,y=7.5,size=1,label=peg-insert-side,shape=ellipse,style={minimum width=70}]{peg-insert-side}
        \Vertex[x=1.5,y=9,size=1,label=push,shape=ellipse,shape=ellipse,style={minimum width=70}]{push}
        \Vertex[x=4.5,y=7.5,size=1,label=window-open,shape=ellipse,style={minimum width=70}]{window-open}
        \Vertex[x=1.5,y=6,size=1,label=drawer-open,shape=ellipse,style={minimum width=70}]{drawer-open}
        \Vertex[x=7.5,y=9,size=1,label=button-press-topdown,shape=ellipse,style={minimum width=70}]{button-press-topdown}
        \Vertex[x=4.5,y=6,size=1,label=door-open,shape=ellipse,style={minimum width=70}]{door-open}
        \Vertex[x=7.5,y=6,size=1,label=drawer-close,shape=ellipse,style={minimum width=70}]{drawer-close}
        \Vertex[x=7.5,y=7.5,size=1,label=window-close,shape=ellipse,style={minimum width=70}]{window-close}
        \Vertex[x=4.5,y=9,size=1,label=reach,shape=ellipse,style={minimum width=70}]{reach}
        \Edge[,Direct](scratch)(pick-place)
        \Edge[,Direct](scratch)(button-press-topdown)
        \Edge[,Direct](pick-place)(push)
        \Edge[,Direct](pick-place)(reach)
        \Edge[,Direct](push)(peg-insert-side)
        \Edge[,Direct](button-press-topdown)(door-open)
        \Edge[,Direct](button-press-topdown)(window-open)
        \Edge[,Direct](button-press-topdown)(window-close)
        \Edge[,Direct](window-open)(drawer-open)
        \Edge[,Direct](window-close)(drawer-close)
      \end{tikzpicture}
    }% This comment prevents a space that would make these pictures not-adjacent
    \scalebox{0.9}{
      \begin{tikzpicture}
        \clip (0,1.5) rectangle (9,11.5);
        \Vertex[x=7.5,y=10.5,size=1.0,label=scratch,shape=rectangle]{scratch}
        \Vertex[x=7.5,y=9,size=1.0,label=drawer-close,shape=ellipse,style={minimum width=70}]{drawer-close}
        \Vertex[x=7.5,y=7.5,size=1.0,label=peg-insert-side,shape=ellipse,style={minimum width=70}]{peg-insert-side}
        \Vertex[x=7.5,y=6,size=1.0,label=button-press-topdown,shape=ellipse,style={minimum width=70}]{button-press-topdown}
        \Vertex[x=7.5,y=4.5,size=1.0,label=drawer-open,shape=ellipse,style={minimum width=70}]{drawer-open}
        \Vertex[x=4.5,y=4.5,size=1.0,label=push,shape=ellipse,style={minimum width=70}]{push}
        \Vertex[x=7.5,y=3,size=1.0,label=window-close,shape=ellipse,style={minimum width=70}]{window-close}
        \Vertex[x=1.5,y=3,size=1.0,label=pick-place,shape=ellipse,style={minimum width=70}]{pick-place}

        \Vertex[x=4.5,y=6,size=1.0,label=door-open,shape=ellipse,style={minimum width=70}]{door-open}
        \Vertex[x=4.5,y=3,size=1.0,label=reach,shape=ellipse,style={minimum width=70}]{reach}
        \Vertex[x=1.5,y=4.5,size=1.0,label=window-open,shape=ellipse,style={minimum width=70}]{window-open}

        \Edge[,Direct](scratch)(drawer-close)
        \Edge[,Direct](drawer-close)(peg-insert-side)
        \Edge[,Direct](push)(reach)
        \Edge[,Direct](peg-insert-side)(button-press-topdown)
        \Edge[,Direct](peg-insert-side)(door-open)
        \Edge[,Direct](button-press-topdown)(drawer-open)
        \Edge[,Direct](door-open)(push)
        \Edge[,Direct](door-open)(pick-place)
        \Edge[,Direct](door-open)(window-open)
        \Edge[,Direct](drawer-open)(window-close)
      \end{tikzpicture}
    }
    \caption{\textbf{Left:} Predicted optimal curriculum DMST for MT10.
    Note that the curriculum begins by learning the \textit{hardest} tasks first, then transfers those skills to easier tasks. \textbf{Right:} Predicted pessimal curriculum DMST for MT10.
  }
    \label{fig:curriculum_trees}
    \push
\end{figure}
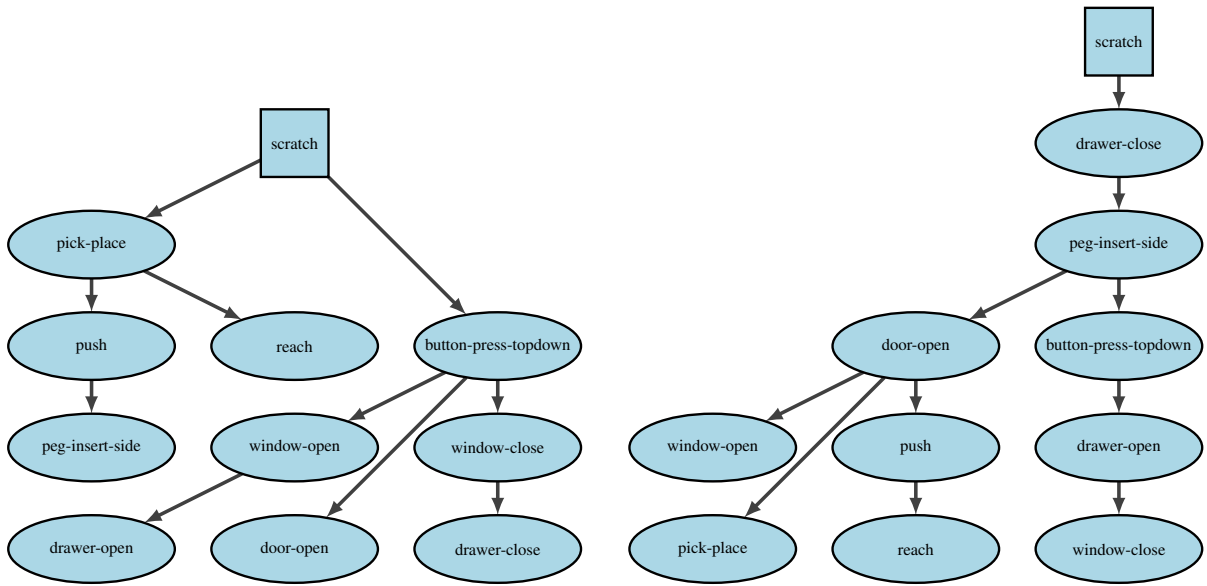

In addition to solving for the DMST to find the optimal curriculum predicted by our cost metric, we can also solve for the Directed Maximum Spanning Tree to find a predicted pessimal curriculum.
See Figure~\ref{fig:curriculum_trees} for examples of optimal and pessimal curriculum trees computed using our DMST-based method for Meta-World MT10, using the skill-skill transfer cost data in Figure~\ref{fig:mt10_matrix}.

Our experience indicates that these graphs make intuitive sense, and tell us a few things about the structure of the task space and the most efficient tasks with which to pre-train for continual learning.
For instance, the optimal curriculum features two distinct sub-trees, with roots corresponding to the two rows of Figure~\ref{fig:mt10_matrix} that contain the largest number of useful transfers ($A_{base \to target} < 1$).

The first sub-tree contains all tasks which can be solved by grasping an object (and reach, which is equivalent to grasping a ``fictional'' object in midair).
It begins with \texttt{pick-place}, which is empirically the most-challenging task in the benchmark, and contains the grasp skill while covering as much of the state space as possible.
It then transfer \texttt{pick-place} to \texttt{push}, which uses the grasp skill to move the grasp object along the table surface, and \texttt{reach}, which it solves trivially.
The second sub-tree contains all other tasks, all of which involve manipulating some object which is fixed to the table across from the table and below the robot gripper's initial position.
In this subtree, \texttt{window-open} is transferred to \texttt{drawer-open} and \texttt{window-close} is transferred \texttt{drawer-close}.
This is somewhat surprising, since one might expect transfer to happen most readily between tasks with the same objects and therefore most similar state distributions.
However, this structure instead corresponds to transferring from more challenging tasks to easier ones with similar motions.

% begins with \texttt{peg-insert-side}, empirically the most challenging task in the benchmark and \texttt{pick-place}, which is empirically the second most-challenging task in the benchmark.
% The optimal curriculum is an approximate ordering of the tasks in the benchmark from last to most challenging, eventually terminating in solving \texttt{reach}, the easiest task.
% The optimal tree branches at \texttt{button-press-topdown} because that skill requires reaching to a location and pushing the gripper down.
% Its terminal child node \texttt{door-open} can be achieved by using the \texttt{button-press-topdown} skill, then simply pulling the robot arm backwards.
% Similarly, its other terminal child node \texttt{drawer-close} is solved even more easily, but reaching forward to close the drawer.

For the pessimal curriculum, similar observations about inter-task structure hold, but with reversed consequences.
The pessimal curriculum instead begins by learning the empirically-easiest task, \texttt{drawer-close} which can be solved very quickly buy simply reaching forward.
It then solves one of the hardest tasks, \texttt{peg-insert-side}, which requires essentially the same number of samples as scratch training when starting from \texttt{drawer-close}.
The remainder of the curriculum alternates between solving easier tasks and harder tasks, terminating in harder tasks.
Each harder-easier transition destroys behaviors useful for future tasks, making the learning process take as long as possible.

Not only do these results suggest that our pairwise transfer cost metric discovers structure in the task space useful for learning skills, it also suggests a somewhat counter-intuitive result: that the most efficient curriculum begins by learning the hardest tasks first, then using them to solve the easier tasks.
Most curriculum learning methods in RL instead learn the easiest tasks first, then use them to generalize to harder tasks.
These results suggest we should initialize our skill library $\mathcal{M}$ by learning policies for the hardest tasks first.
% KR: I still don't understand the different between transferring behavior vs. transferring parameters.
% They also confirm one of the major hypotheses: that transferring policy behavior, not parameters, is most important for efficient continual manipulation skill learning.
We empirically confirm these results with continual learning experiments using these curricula.

% Once we have a curriculum tree, we can use it to learn all tasks by training skills for all tasks directly connected from the ``scratch'' vertex, then solve the remaining tasks in the tree by repeatedly picking an unsolved task $\tau'$ connected to a solved task $\tau'$ by an edge, and solving $\tau'$ by transferring skills from $\tau$.
% We refer to this algorithm as Iterated Skill Transfer.
% \todo{It would be nice to have an algorithm figure for Iterated Skill Transfer}

Given the solved optimal curriculum tree $T_{optimal}$, Curriculum Skill Transfer learns all tasks continually by learning skill in the sequence produced by the tree traversal on $T_{optimal}$, starting from the root ``scratch'' node (Algorithm~\ref{alg:curriculum_finetuning}).
Note that if we reject a transfer that was predicted to succeed by our cost metric, we remove that edge from the graph, re-compute the curriculum using DMST to find a better curriculum online, and continue learning under the new curriculum.

\hspace*{-\parindent}%
\begin{minipage}{\columnwidth}
\begin{algorithm}[H]
    \begin{algorithmic}[1]
      \STATE {\bfseries Input:} Initial skill library $\mathcal{M}_0$, target task space $\mathcal{T}$, RL algorithm $\mathcal{F} \to (\pi, C)$
      \STATE $V \gets \mathcal{T} \cup scratch$
      \STATE $E \gets \{\}$
      \FOR{$\tau_{base} \in \mathcal{T}$}
        \STATE $E \gets (scratch, \tau_{base}, -1.0)$
        \STATE $\pi_{base}, C_{scratch \to target} \gets \mathcal{F}(\tau_{base}, \pi_{random})$
        \FOR{$\tau_{target} \in \mathcal{T}$}
          \STATE $\cdot, C_{base \to target} = \mathcal{F}(\tau_{target}, \pi_{base})$
          \STATE $E \gets E (\tau_{base}, \tau_{target}, C_{base \to target}) \cup E$
        \ENDFOR
      \ENDFOR
      \STATE $T_{optimal} \gets \mathtt{kruskal}((V, E))$
      \STATE $i \gets 1$
      \STATE $\pi_{base} \gets \pi_{random}$
      \FOR{$\tau_{target} \in \mathtt{traverse}(T_{optimal})$}
        \WHILE{$\tau_{target}$ not solved}
          \STATE $\pi_{target}, \cdot \gets \mathcal{F}(\tau_{target}, \mathtt{clone}(\pi_{base}))$
          \IF{$\tau_{target}$ not solved}
            \STATE $E \gets E \setminus (\tau_{base}, \tau_{target})$
            \STATE $T_{optimal} \gets \mathtt{kruskal}((V, E))$
          \ENDIF
        \ENDWHILE
        \STATE $\mathcal{M}_{i} \gets \{\pi_{target}\}\cup \mathcal{M}_{i-1}$
        \STATE $i \gets i + 1$
      \ENDFOR
      \STATE {\bfseries Output:} Skill library $\mathcal{M}$
    \end{algorithmic}
    \caption{DMST-Based Curriculum Transfer}
   \label{alg:curriculum_finetuning}
\end{algorithm}
\vspace{0.5em}
\end{minipage}

\subsection{Measuring the Effectiveness of Curriculum Selection}
\begin{figure}[h]
    \centering
    \includegraphics[width=\columnwidth]{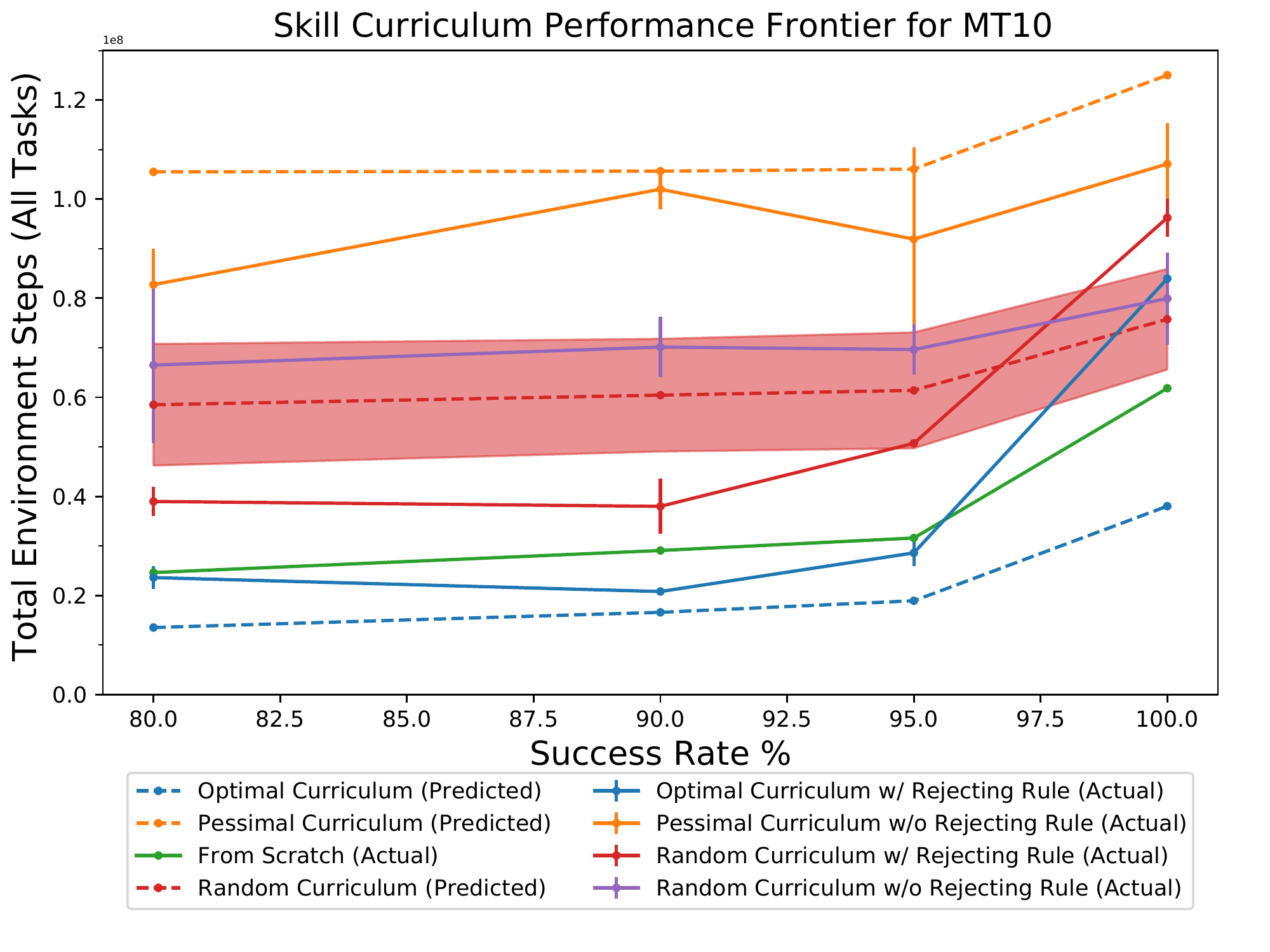}
    \caption{Comparison of the performance-sample efficiency frontier for several curricula for Meta-World MT10.
    The high level of agreement between the actual and predicted curricula indicate that our skill-skill transfer cost metric accurately predicts the true cost to transfer skills.
    Note that the success rate is presented on the X-axis, since it is chosen as the point from which to stop learning one task and begin learning another.
    Error bars show one standard deviation in number of total environment steps require to solve all tasks.
    However, runs using the rejecting rule exhibit very low variance between runs.
    }
    \label{fig:curriculum_frontier}
\end{figure}

We demonstrate the effectiveness of our curriculum selection algorithm in Figure~\ref{fig:curriculum_frontier}, in which the total cost in environment steps is shown for each success rate we could use as a success criteria, calculated using the data from the skill-skill cost metric training experiments.
The optimal curriculum computed using our curriculum selection algorithm matches or outperforms training from scratch for success rates between $80\%$ and $95\%$.

% uses fewer samples than training from scratch at all performance criteria.

% Both pessimal and random curricula are far less sample efficient than our predicted optimal curriculum or training from scratch.
% Curriculum Skill Transfer significantly outperforms training from scratch at $90\%$ success.
% Note that we show confidence intervals for runs of our optimal curriculum.
% However, because of our restarting rule and the large total time for the continual learning run, the variance between runs is very small.

%Besides predicting the total cost of a curriculum using the skill-skill experiments., we can apply our cost metric to an actual continual learning experiment, to compute a predicted sample efficiency

%This may change the edges in the curriculum, so we do not provide predicted values for success criteria we did not run.

%% file: 07_conclusion.tex
In this work, we introduced a simple approach to continual learning for robotic manipulation, based around a growing library of skill policies and repeated skill-skill fine-tuning.
% We first informed the intuition behind Skill Builder's design with simple skill policy adaptation experiments, which show that both skill policy parameters and behavior are essential for efficient policy adaptation.
% We first introduced the method as an abstract framework, and showed that a na{\"i}ve implementation of that framework (Random Skill Transfer) achieves continual skill learning without forgetting, but is not more efficient than training all skills from scratch.
We first introduced the method as an abstract framework, and showed that a na{\"i}ve implementation of that framework (Random Skill Transfer) achieves continual skill learning without forgetting, but is not more efficient than training all skills from scratch.
We discussed the importance of skill curricula for efficient continual learning, reduced the efficient continual learning problem to minimizing total skill-skill transfer cost, developed an offline metric for measuring that cost based on skill-skill fine-tuning performance, and used this metric to solve for optimal and pessimal curricula using an DMST-based algorithm.
We found that the optimal and pessimal curricula produced by this DMST-based curriculum algorithm are not only intuitive, but they make clear a rather unintuitive result: that it is best to pre-train continual learners with the hardest manipulation skills first.
Finally, we tested these curricula to continually learn Meta-World MT10, and verified that continual learning with the optimal curriculum outperforms training from scratch, and the pessimal curriculum performs much worse than both training scratch and a random curriculum.
% Finally, we introduced a policy model class for active online curriculum selection during fine-tuning, and shared promising results towards using it for continual learning.

In future work, we will investigate ways of approximating our cost metric, as well as policy model classes that can transfer skills without explicit cost metrics.
We look forward to developing a method for active online curriculum selection, and using it to achieve efficient continual learning.

% skill-skill transfer cost metrics which can be inferred online
% We will also investigate different skill-skill transfer cost metrics, especially those which might be inferred online, or by learning an inference model during pre-training which predicts the cost of learning future skills given the pre-train tasks.

%% file: 00_main.bbl
\begin{thebibliography}{134}
\providecommand{\natexlab}[1]{#1}
\providecommand{\url}[1]{\texttt{#1}}
\expandafter\ifx\csname urlstyle\endcsname\relax
  \providecommand{\doi}[1]{doi: #1}\else
  \providecommand{\doi}{doi: \begingroup \urlstyle{rm}\Url}\fi

\bibitem[Mnih et~al.(2013)Mnih, Kavukcuoglu, Silver, Graves, Antonoglou,
  Wierstra, and Riedmiller]{mnih2013playing}
Volodymyr Mnih, Koray Kavukcuoglu, David Silver, Alex Graves, Ioannis
  Antonoglou, Daan Wierstra, and Martin Riedmiller.
\newblock Playing atari with deep reinforcement learning.
\newblock \emph{arXiv preprint arXiv:1312.5602}, 2013.

\bibitem[Mnih et~al.(2015)Mnih, Kavukcuoglu, Silver, Rusu, Veness, Bellemare,
  Graves, Riedmiller, Fidjeland, Ostrovski, et~al.]{mnih2015human}
Volodymyr Mnih, Koray Kavukcuoglu, David Silver, Andrei~A Rusu, Joel Veness,
  Marc~G Bellemare, Alex Graves, Martin Riedmiller, Andreas~K Fidjeland, Georg
  Ostrovski, et~al.
\newblock Human-level control through deep reinforcement learning.
\newblock \emph{Nature}, 518\penalty0 (7540):\penalty0 529--533, 2015.
\newblock ISSN 14764687.
\newblock \doi{10.1038/nature14236}.
\newblock URL
  \url{https://www.nature.com/nature/journal/v518/n7540/pdf/nature1423\\6.pdf}.

\bibitem[Bellemare et~al.(2020)Bellemare, Candido, Castro, Gong, Machado,
  Moitra, Ponda, and Wang]{bellemare2020autonomous}
Marc~G Bellemare, Salvatore Candido, Pablo~Samuel Castro, Jun Gong, Marlos~C
  Machado, Subhodeep Moitra, Sameera~S Ponda, and Ziyu Wang.
\newblock Autonomous navigation of stratospheric balloons using reinforcement
  learning.
\newblock \emph{Nature}, 588\penalty0 (7836):\penalty0 77--82, 2020.

\bibitem[Levine et~al.(2016)Levine, Finn, Darrell, and Abbeel]{levine2016end}
Sergey Levine, Chelsea Finn, Trevor Darrell, and Pieter Abbeel.
\newblock End-to-end training of deep visuomotor policies.
\newblock \emph{The Journal of Machine Learning Research}, 17\penalty0
  (1):\penalty0 1334--1373, 2016.

\bibitem[Julian et~al.(2020)Julian, Swanson, Sukhatme, Levine, Finn, and
  Hausman]{julian2020never}
Ryan Julian, Benjamin Swanson, Gaurav~S Sukhatme, Sergey Levine, Chelsea Finn,
  and Karol Hausman.
\newblock Never stop learning: The effectiveness of fine-tuning in robotic
  reinforcement learning.
\newblock \emph{arXiv e-prints}, pages arXiv--2004, 2020.

\bibitem[Schulman et~al.(2017)Schulman, Wolski, Dhariwal, Radford, and
  Klimov]{schulman2017proximal}
John Schulman, Filip Wolski, Prafulla Dhariwal, Alec Radford, and Oleg Klimov.
\newblock Proximal policy optimization algorithms.
\newblock \emph{arXiv preprint arXiv:1707.06347}, 2017.

\bibitem[Kober et~al.(2013)Kober, Bagnell, and Peters]{kober2013reinforcement}
Jens Kober, J~Andrew Bagnell, and Jan Peters.
\newblock Reinforcement learning in robotics: A survey.
\newblock \emph{The International Journal of Robotics Research}, 32\penalty0
  (11):\penalty0 1238--1274, 2013.

\bibitem[Mahadevan and Connell(1992)]{mahadevan1992automatic}
Sridhar Mahadevan and Jonathan Connell.
\newblock Automatic programming of behavior-based robots using reinforcement
  learning.
\newblock \emph{Artificial intelligence}, 55\penalty0 (2-3):\penalty0 311--365,
  1992.

\bibitem[Lin(1992)]{lin1992reinforcement}
Long-Ji Lin.
\newblock Reinforcement learning for robots using neural networks, 1992.

\bibitem[Smart and Kaelbling(2002)]{smart2002effective}
William~D Smart and L~Pack Kaelbling.
\newblock Effective reinforcement learning for mobile robots.
\newblock In \emph{Proceedings 2002 IEEE International Conference on Robotics
  and Automation (Cat. No. 02CH37292)}, volume~4, pages 3404--3410. IEEE, 2002.

\bibitem[Levine et~al.(2018)Levine, Pastor, Krizhevsky, Ibarz, and
  Quillen]{levine2018robotarmy}
Sergey Levine, Peter Pastor, Alex Krizhevsky, Julian Ibarz, and Deirdre
  Quillen.
\newblock Learning hand-eye coordination for robotic grasping with deep
  learning and large-scale data collection.
\newblock \emph{The International Journal of Robotics Research}, 37\penalty0
  (4-5):\penalty0 421--436, 2018.
\newblock \doi{10.1177/0278364917710318}.
\newblock URL \url{https://doi.org/10.1177/0278364917710318}.

\bibitem[Kalashnikov et~al.(2018)Kalashnikov, Irpan, Pastor, Ibarz, Herzog,
  Jang, Quillen, Holly, Kalakrishnan, Vanhoucke,
  et~al.]{kalashnikov2018scalable}
Dmitry Kalashnikov, Alex Irpan, Peter Pastor, Julian Ibarz, Alexander Herzog,
  Eric Jang, Deirdre Quillen, Ethan Holly, Mrinal Kalakrishnan, Vincent
  Vanhoucke, et~al.
\newblock Scalable deep reinforcement learning for vision-based robotic
  manipulation.
\newblock In \emph{Conference on Robot Learning}, pages 651--673, 2018.

\bibitem[Pinto and Gupta(2016)]{pinto2016supersizing}
Lerrel Pinto and Abhinav Gupta.
\newblock Supersizing self-supervision: Learning to grasp from 50k tries and
  700 robot hours.
\newblock In \emph{2016 IEEE international conference on robotics and
  automation (ICRA)}, pages 3406--3413. IEEE, 2016.

\bibitem[Gullapalli et~al.(1994)Gullapalli, Franklin, and
  Benbrahim]{gullapalli1994acquiring}
Vijaykumar Gullapalli, Judy~A Franklin, and Hamid Benbrahim.
\newblock Acquiring robot skills via reinforcement learning.
\newblock \emph{IEEE Control Systems Magazine}, 14\penalty0 (1):\penalty0
  13--24, 1994.

\bibitem[Ghadirzadeh et~al.(2017)Ghadirzadeh, Maki, Kragic, and
  Bj{\"o}rkman]{ghadirzadeh2017deep}
Ali Ghadirzadeh, Atsuto Maki, Danica Kragic, and M{\aa}rten Bj{\"o}rkman.
\newblock Deep predictive policy training using reinforcement learning.
\newblock In \emph{2017 IEEE/RSJ International Conference on Intelligent Robots
  and Systems (IROS)}, pages 2351--2358. IEEE, 2017.

\bibitem[Zeng et~al.(2018)Zeng, Song, Welker, Lee, Rodriguez, and
  Funkhouser]{zeng2018learning}
Andy Zeng, Shuran Song, Stefan Welker, Johnny Lee, Alberto Rodriguez, and
  Thomas Funkhouser.
\newblock Learning synergies between pushing and grasping with self-supervised
  deep reinforcement learning.
\newblock \emph{arXiv preprint arXiv:1803.09956}, 2018.

\bibitem[Kohl and Stone(2004{\natexlab{a}})]{kohl2004policy}
Nate Kohl and Peter Stone.
\newblock Policy gradient reinforcement learning for fast quadrupedal
  locomotion.
\newblock In \emph{IEEE International Conference on Robotics and Automation,
  2004. Proceedings. ICRA'04. 2004}, volume~3, pages 2619--2624. IEEE,
  2004{\natexlab{a}}.

\bibitem[Kohl and Stone(2004{\natexlab{b}})]{stone2004machine}
Nate Kohl and Peter Stone.
\newblock Machine learning for fast quadrupedal locomotion.
\newblock In \emph{The Nineteenth National Conference on Artificial
  Intelligence}, pages 611--616, July 2004{\natexlab{b}}.

\bibitem[Xie et~al.(2019)Xie, Clary, Dao, Morais, Hurst, and van~de
  Panne]{xie2019iterative}
Zhaoming Xie, Patrick Clary, Jeremy Dao, Pedro Morais, Jonathan Hurst, and
  Michiel van~de Panne.
\newblock Iterative reinforcement learning based design of dynamic locomotion
  skills for cassie.
\newblock \emph{arXiv preprint arXiv:1903.09537}, 2019.

\bibitem[Haarnoja et~al.(2019)Haarnoja, Ha, Zhou, Tan, Tucker, and
  Levine]{haarnoja2019learning}
Tuomas Haarnoja, Sehoon Ha, Aurick Zhou, Jie Tan, George Tucker, and Sergey
  Levine.
\newblock Learning to walk via deep reinforcement learning.
\newblock In \emph{Robotics: Science and Systems}, 2019.

\bibitem[Beom and Cho(1995)]{beom1995sensor}
Hee~Rak Beom and Hyung~Suck Cho.
\newblock A sensor-based navigation for a mobile robot using fuzzy logic and
  reinforcement learning.
\newblock \emph{IEEE transactions on Systems, Man, and Cybernetics},
  25\penalty0 (3):\penalty0 464--477, 1995.

\bibitem[Zhu et~al.(2017)Zhu, Mottaghi, Kolve, Lim, Gupta, Fei-Fei, and
  Farhadi]{zhu2017target}
Yuke Zhu, Roozbeh Mottaghi, Eric Kolve, Joseph~J Lim, Abhinav Gupta,
  Li~Fei-Fei, and Ali Farhadi.
\newblock Target-driven visual navigation in indoor scenes using deep
  reinforcement learning.
\newblock In \emph{2017 IEEE international conference on robotics and
  automation (ICRA)}, pages 3357--3364. IEEE, 2017.

\bibitem[Singh et~al.(1994)Singh, Barto, Grupen, Connolly,
  et~al.]{singh1994robust}
Satinder~P Singh, Andrew~G Barto, Roderic Grupen, Christopher Connolly, et~al.
\newblock Robust reinforcement learning in motion planning.
\newblock \emph{Advances in neural information processing systems}, pages
  655--655, 1994.

\bibitem[Everett et~al.(2018)Everett, Chen, and How]{everett2018motion}
Michael Everett, Yu~Fan Chen, and Jonathan~P How.
\newblock Motion planning among dynamic, decision-making agents with deep
  reinforcement learning.
\newblock In \emph{2018 IEEE/RSJ International Conference on Intelligent Robots
  and Systems (IROS)}, pages 3052--3059. IEEE, 2018.

\bibitem[Bagnell and Schneider(2001)]{bagnell2001autonomous}
J~Andrew Bagnell and Jeff~G Schneider.
\newblock Autonomous helicopter control using reinforcement learning policy
  search methods.
\newblock In \emph{Proceedings 2001 ICRA. IEEE International Conference on
  Robotics and Automation (Cat. No. 01CH37164)}, volume~2, pages 1615--1620.
  IEEE, 2001.

\bibitem[Abbeel et~al.(2007)Abbeel, Coates, Quigley, and
  Ng]{abbeel2007application}
Pieter Abbeel, Adam Coates, Morgan Quigley, and Andrew~Y Ng.
\newblock An application of reinforcement learning to aerobatic helicopter
  flight.
\newblock \emph{Advances in neural information processing systems},
  19:\penalty0 1, 2007.

\bibitem[Ng et~al.(2003)Ng, Kim, Jordan, and Sastry]{ng2003autonomous}
Andrew~Y Ng, H~Jin Kim, Michael~I Jordan, and Shankar Sastry.
\newblock Autonomous helicopter flight via reinforcement learning.
\newblock In \emph{Proceedings of the 16th International Conference on Neural
  Information Processing Systems}, pages 799--806, 2003.

\bibitem[Matari{\'c}(1997)]{mataric1997reinforcement}
Maja~J Matari{\'c}.
\newblock Reinforcement learning in the multi-robot domain.
\newblock \emph{Autonomous Robots}, 4\penalty0 (1):\penalty0 73--83, 1997.

\bibitem[Yang and Gu(2004)]{yang2004multiagent}
Erfu Yang and Dongbing Gu.
\newblock Multiagent reinforcement learning for multi-robot systems: A survey.
\newblock Technical report, Tech Report, 2004.

\bibitem[Long et~al.(2018)Long, Fan, Liao, Liu, Zhang, and
  Pan]{long2018towards}
Pinxin Long, Tingxiang Fan, Xinyi Liao, Wenxi Liu, Hao Zhang, and Jia Pan.
\newblock Towards optimally decentralized multi-robot collision avoidance via
  deep reinforcement learning.
\newblock In \emph{2018 IEEE International Conference on Robotics and
  Automation (ICRA)}, pages 6252--6259. IEEE, 2018.

\bibitem[Bengio et~al.(2017)Bengio, Goodfellow, and Courville]{bengio2017deep}
Yoshua Bengio, Ian Goodfellow, and Aaron Courville.
\newblock \emph{Deep learning}, volume~1.
\newblock MIT press Massachusetts, USA:, 2017.

\bibitem[Fran{\c{c}}ois-Lavet et~al.(2018)Fran{\c{c}}ois-Lavet, Henderson,
  Islam, Bellemare, and Pineau]{franccois2018introduction}
Vincent Fran{\c{c}}ois-Lavet, Peter Henderson, Riashat Islam, Marc~G Bellemare,
  and Joelle Pineau.
\newblock An introduction to deep reinforcement learning.
\newblock \emph{arXiv preprint arXiv:1811.12560}, 2018.

\bibitem[Lillicrap et~al.(2015)Lillicrap, Hunt, Pritzel, Heess, Erez, Tassa,
  Silver, and Wierstra]{lillicrap2015continuous}
Timothy~P Lillicrap, Jonathan~J Hunt, Alexander Pritzel, Nicolas Heess, Tom
  Erez, Yuval Tassa, David Silver, and Daan Wierstra.
\newblock Continuous control with deep reinforcement learning.
\newblock \emph{arXiv preprint arXiv:1509.02971}, 2015.

\bibitem[Hadsell et~al.(2009)Hadsell, Sermanet, Ben, Erkan, Scoffier,
  Kavukcuoglu, Muller, and LeCun]{hadsell2009learning}
Raia Hadsell, Pierre Sermanet, Jan Ben, Ayse Erkan, Marco Scoffier, Koray
  Kavukcuoglu, Urs Muller, and Yann LeCun.
\newblock Learning long-range vision for autonomous off-road driving.
\newblock \emph{Journal of Field Robotics}, 26\penalty0 (2):\penalty0 120--144,
  2009.

\bibitem[Donahue et~al.(2014)Donahue, Jia, Vinyals, Hoffman, Zhang, Tzeng, and
  Darrell]{donahue2014decaf}
Jeff Donahue, Yangqing Jia, Oriol Vinyals, Judy Hoffman, Ning Zhang, Eric
  Tzeng, and Trevor Darrell.
\newblock Decaf: A deep convolutional activation feature for generic visual
  recognition.
\newblock In \emph{International conference on machine learning}, pages
  647--655, 2014.

\bibitem[Howard and Ruder(2018)]{ulmfit}
Jeremy Howard and Sebastian Ruder.
\newblock Universal language model fine-tuning for text classification, 2018.

\bibitem[Devlin et~al.(2018)Devlin, Chang, Lee, and Toutanova]{devlin2018bert}
Jacob Devlin, Ming-Wei Chang, Kenton Lee, and Kristina Toutanova.
\newblock Bert: Pre-training of deep bidirectional transformers for language
  understanding.
\newblock \emph{arXiv preprint arXiv:1810.04805}, 2018.

\bibitem[Dai et~al.(2007)Dai, Yang, Xue, and Yu]{dai2007boosting}
Wenyuan Dai, Qiang Yang, Gui-Rong Xue, and Yong Yu.
\newblock Boosting for transfer learning.
\newblock In \emph{Proceedings of the 24th international conference on Machine
  learning}, pages 193--200, 2007.

\bibitem[Raina et~al.(2007)Raina, Battle, Lee, Packer, and Ng]{raina2007self}
Rajat Raina, Alexis Battle, Honglak Lee, Benjamin Packer, and Andrew~Y Ng.
\newblock Self-taught learning: transfer learning from unlabeled data.
\newblock In \emph{Proceedings of the 24th international conference on Machine
  learning}, pages 759--766, 2007.

\bibitem[Silver et~al.(2018)Silver, Allen, Tenenbaum, and
  Kaelbling]{silver2018residual}
Tom Silver, Kelsey Allen, Josh Tenenbaum, and Leslie Kaelbling.
\newblock Residual policy learning.
\newblock \emph{arXiv preprint arXiv:1812.06298}, 2018.

\bibitem[Ruder(2017)]{ruder2017overview}
Sebastian Ruder.
\newblock An overview of multi-task learning in deep neural networks, 2017.

\bibitem[Rusu et~al.(2016{\natexlab{a}})Rusu, Rabinowitz, Desjardins, Soyer,
  Kirkpatrick, Kavukcuoglu, Pascanu, and Hadsell]{rusu2016progressive}
Andrei~A Rusu, Neil~C Rabinowitz, Guillaume Desjardins, Hubert Soyer, James
  Kirkpatrick, Koray Kavukcuoglu, Razvan Pascanu, and Raia Hadsell.
\newblock Progressive neural networks.
\newblock \emph{arXiv preprint arXiv:1606.04671}, abs/1606.04671,
  2016{\natexlab{a}}.
\newblock URL \url{http://arxiv.org/abs/1606.04671}.

\bibitem[Finn and Levine(2017)]{finn2017deep}
Chelsea Finn and Sergey Levine.
\newblock Deep visual foresight for planning robot motion.
\newblock In \emph{2017 IEEE International Conference on Robotics and
  Automation (ICRA)}, pages 2786--2793. IEEE, 2017.

\bibitem[Yen-Chen et~al.(2019)Yen-Chen, Bauza, and Isola]{yen2019experience}
Lin Yen-Chen, Maria Bauza, and Phillip Isola.
\newblock Experience-embedded visual foresight.
\newblock \emph{arXiv preprint arXiv:1911.05071}, 2019.

\bibitem[Nagabandi et~al.(2019)Nagabandi, Konoglie, Levine, and
  Kumar]{nagabandi2019deep}
Anusha Nagabandi, Kurt Konoglie, Sergey Levine, and Vikash Kumar.
\newblock Deep dynamics models for learning dexterous manipulation.
\newblock \emph{arXiv preprint arXiv:1909.11652}, 2019.

\bibitem[Chatzilygeroudis and Mouret(2018)]{chatzilygeroudis2018using}
Konstantinos Chatzilygeroudis and Jean-Baptiste Mouret.
\newblock Using parameterized black-box priors to scale up model-based policy
  search for robotics.
\newblock In \emph{2018 IEEE International Conference on Robotics and
  Automation (ICRA)}, pages 1--9. IEEE, 2018.

\bibitem[Ha and Schmidhuber(2018)]{ha2018recurrent}
David Ha and J{\"u}rgen Schmidhuber.
\newblock Recurrent world models facilitate policy evolution.
\newblock In \emph{Advances in Neural Information Processing Systems}, pages
  2450--2462, 2018.

\bibitem[Dasari et~al.(2019)Dasari, Ebert, Tian, Nair, Bucher, Schmeckpeper,
  Singh, Levine, and Finn]{dasari2019robonet}
Sudeep Dasari, Frederik Ebert, Stephen Tian, Suraj Nair, Bernadette Bucher,
  Karl Schmeckpeper, Siddharth Singh, Sergey Levine, and Chelsea Finn.
\newblock Robonet: Large-scale multi-robot learning, 2019.

\bibitem[Chatzilygeroudis et~al.(2018)Chatzilygeroudis, Vassiliades, and
  Mouret]{chatzilygeroudis2018reset}
Konstantinos Chatzilygeroudis, Vassilis Vassiliades, and Jean-Baptiste Mouret.
\newblock Reset-free trial-and-error learning for robot damage recovery.
\newblock \emph{Robotics and Autonomous Systems}, 100:\penalty0 236--250, 2018.

\bibitem[Cully et~al.(2015)Cully, Clune, Tarapore, and Mouret]{cully2015robots}
Antoine Cully, Jeff Clune, Danesh Tarapore, and Jean-Baptiste Mouret.
\newblock Robots that can adapt like animals.
\newblock \emph{Nature}, 521\penalty0 (7553):\penalty0 503--507, 2015.

\bibitem[Kaushik et~al.(2020)Kaushik, Desreumaux, and
  Mouret]{kaushik2020adaptive}
Rituraj Kaushik, Pierre Desreumaux, and Jean-Baptiste Mouret.
\newblock Adaptive prior selection for repertoire-based online adaptation in
  robotics.
\newblock \emph{Frontiers in Robotics and AI}, 6:\penalty0 151, 2020.

\bibitem[Merel et~al.(2019)Merel, Tunyasuvunakool, Ahuja, Tassa, Hasenclever,
  Pham, Erez, Wayne, and Heess]{merel2019reusable}
Josh Merel, Saran Tunyasuvunakool, Arun Ahuja, Yuval Tassa, Leonard
  Hasenclever, Vu~Pham, Tom Erez, Greg Wayne, and Nicolas Heess.
\newblock Reusable neural skill embeddings for vision-guided whole body
  movement and object manipulation.
\newblock \emph{arXiv preprint arXiv:1911.06636}, 2019.

\bibitem[Rastogi et~al.(2018)Rastogi, Koryakovskiy, and
  Kober]{rastogi2018sample}
Divyam Rastogi, Ivan Koryakovskiy, and Jens Kober.
\newblock Sample-efficient reinforcement learning via difference models.
\newblock In \emph{Technical Report}, 2018.

\bibitem[Jeong et~al.(2019)Jeong, Kay, Romano, Lampe, Rothorl, Abdolmaleki,
  Erez, Tassa, and Nori]{jeong2019modelling}
Rae Jeong, Jackie Kay, Francesco Romano, Thomas Lampe, Tom Rothorl, Abbas
  Abdolmaleki, Tom Erez, Yuval Tassa, and Francesco Nori.
\newblock Modelling generalized forces with reinforcement learning for
  sim-to-real transfer.
\newblock \emph{arXiv preprint arXiv:1910.09471}, 2019.

\bibitem[Agrawal et~al.(2016)Agrawal, Nair, Abbeel, Malik, and
  Levine]{agrawal2016learning}
Pulkit Agrawal, Ashvin~V Nair, Pieter Abbeel, Jitendra Malik, and Sergey
  Levine.
\newblock Learning to poke by poking: Experiential learning of intuitive
  physics.
\newblock In \emph{Advances in neural information processing systems}, pages
  5074--5082, 2016.

\bibitem[Nair et~al.(2018)Nair, Pong, Dalal, Bahl, Lin, and
  Levine]{nair2018visual}
Ashvin~V Nair, Vitchyr Pong, Murtaza Dalal, Shikhar Bahl, Steven Lin, and
  Sergey Levine.
\newblock Visual reinforcement learning with imagined goals.
\newblock In \emph{Advances in Neural Information Processing Systems}, pages
  9191--9200, 2018.

\bibitem[Pathak et~al.(2018)Pathak, Mahmoudieh, Luo, Agrawal, Chen, Shentu,
  Shelhamer, Malik, Efros, and Darrell]{Pathak_2018}
Deepak Pathak, Parsa Mahmoudieh, Guanghao Luo, Pulkit Agrawal, Dian Chen, Fred
  Shentu, Evan Shelhamer, Jitendra Malik, Alexei~A. Efros, and Trevor Darrell.
\newblock Zero-shot visual imitation.
\newblock \emph{2018 IEEE/CVF Conference on Computer Vision and Pattern
  Recognition Workshops (CVPRW)}, June 2018.
\newblock \doi{10.1109/cvprw.2018.00278}.
\newblock URL \url{http://dx.doi.org/10.1109/CVPRW.2018.00278}.

\bibitem[Pong et~al.(2019)Pong, Dalal, Lin, Nair, Bahl, and
  Levine]{pong2019skew}
Vitchyr~H Pong, Murtaza Dalal, Steven Lin, Ashvin Nair, Shikhar Bahl, and
  Sergey Levine.
\newblock Skew-fit: State-covering self-supervised reinforcement learning.
\newblock \emph{arXiv preprint arXiv:1903.03698}, 2019.

\bibitem[Yu et~al.(2019{\natexlab{a}})Yu, Shevchuk, Sadigh, and
  Finn]{yu2019unsupervised}
Tianhe Yu, Gleb Shevchuk, Dorsa Sadigh, and Chelsea Finn.
\newblock Unsupervised visuomotor control through distributional planning
  networks.
\newblock \emph{arXiv preprint arXiv:1902.05542}, 2019{\natexlab{a}}.

\bibitem[Nagabandi et~al.(2018{\natexlab{a}})Nagabandi, Clavera, Liu, Fearing,
  Abbeel, Levine, and Finn]{nagabandi2018learning}
Anusha Nagabandi, Ignasi Clavera, Simin Liu, Ronald~S Fearing, Pieter Abbeel,
  Sergey Levine, and Chelsea Finn.
\newblock Learning to adapt in dynamic, real-world environments through
  meta-reinforcement learning.
\newblock \emph{arXiv:1803.11347}, 2018{\natexlab{a}}.
\newblock URL \url{https://openreview.net/forum?id=HyztsoC5Y7}.

\bibitem[Alet et~al.(2018)Alet, Lozano-P{\'e}rez, and
  Kaelbling]{alet2018modular}
Ferran Alet, Tom{\'a}s Lozano-P{\'e}rez, and Leslie~P Kaelbling.
\newblock Modular meta-learning.
\newblock \emph{arXiv preprint arXiv:1806.10166}, 2018.

\bibitem[Nagabandi et~al.(2018{\natexlab{b}})Nagabandi, Finn, and
  Levine]{nagabandi2018deep}
Anusha Nagabandi, Chelsea Finn, and Sergey Levine.
\newblock Deep online learning via meta-learning: Continual adaptation for
  model-based rl.
\newblock \emph{arXiv preprint arXiv:1812.07671}, 2018{\natexlab{b}}.

\bibitem[Finn et~al.(2017)Finn, Yu, Zhang, Abbeel, and Levine]{finn2017one}
Chelsea Finn, Tianhe Yu, Tianhao Zhang, Pieter Abbeel, and Sergey Levine.
\newblock One-shot visual imitation learning via meta-learning.
\newblock \emph{arXiv preprint arXiv:1709.04905}, 2017.

\bibitem[James et~al.(2018)James, Bloesch, and Davison]{james2018task}
Stephen James, Michael Bloesch, and Andrew~J Davison.
\newblock Task-embedded control networks for few-shot imitation learning.
\newblock In \emph{Conference on Robot Learning}, pages 783--795, 2018.

\bibitem[Yu et~al.(2018)Yu, Finn, Xie, Dasari, Zhang, Abbeel, and
  Levine]{yu2018one}
Tianhe Yu, Chelsea Finn, Annie Xie, Sudeep Dasari, Tianhao Zhang, Pieter
  Abbeel, and Sergey Levine.
\newblock One-shot imitation from observing humans via domain-adaptive
  meta-learning.
\newblock In \emph{International Conference on Learning Representations}, 2018.

\bibitem[Bonardi et~al.(2019)Bonardi, James, and Davison]{bonardi2019learning}
Alessandro Bonardi, Stephen James, and Andrew~J Davison.
\newblock Learning one-shot imitation from humans without humans.
\newblock \emph{arXiv preprint arXiv:1911.01103}, 2019.

\bibitem[Yu et~al.(2019{\natexlab{b}})Yu, Quillen, He, Julian, Hausman, Finn,
  and Levine]{yu2019meta}
Tianhe Yu, Deirdre Quillen, Zhanpeng He, Ryan Julian, Karol Hausman, Chelsea
  Finn, and Sergey Levine.
\newblock Meta-world: A benchmark and evaluation for multi-task and meta
  reinforcement learning.
\newblock \emph{arXiv preprint arXiv:1910.10897}, 2019{\natexlab{b}}.

\bibitem[Deng et~al.(2009)Deng, Dong, Socher, Li, Li, and
  Fei-Fei]{deng2009imagenet}
Jia Deng, Wei Dong, Richard Socher, Li-Jia Li, Kai Li, and Li~Fei-Fei.
\newblock Imagenet: A large-scale hierarchical image database.
\newblock In \emph{2009 IEEE conference on computer vision and pattern
  recognition}, pages 248--255. Ieee, 2009.

\bibitem[Finn et~al.(2016)Finn, Tan, Duan, Darrell, Levine, and
  Abbeel]{finn2016deep}
Chelsea Finn, Xin~Yu Tan, Yan Duan, Trevor Darrell, Sergey Levine, and Pieter
  Abbeel.
\newblock Deep spatial autoencoders for visuomotor learning.
\newblock In \emph{2016 IEEE International Conference on Robotics and
  Automation (ICRA)}, pages 512--519. IEEE, 2016.

\bibitem[Gupta et~al.(2018)Gupta, Murali, Gandhi, and Pinto]{gupta2018robot}
Abhinav Gupta, Adithyavairavan Murali, Dhiraj~Prakashchand Gandhi, and Lerrel
  Pinto.
\newblock Robot learning in homes: Improving generalization and reducing
  dataset bias.
\newblock In \emph{Advances in Neural Information Processing Systems}, pages
  9112--9122, 2018.

\bibitem[Sadeghi and Levine(2017)]{sadeghi2017cadrl}
Fereshteh Sadeghi and Sergey Levine.
\newblock Cad2rl: Real single-image flight without a single real image.
\newblock \emph{Robotics: Science and Systems XIII}, July 2017.
\newblock \doi{10.15607/rss.2017.xiii.034}.
\newblock URL \url{http://dx.doi.org/10.15607/RSS.2017.XIII.034}.

\bibitem[Tobin et~al.(2017)Tobin, Fong, Ray, Schneider, Zaremba, and
  Abbeel]{tobin2017domain}
Josh Tobin, Rachel Fong, Alex Ray, Jonas Schneider, Wojciech Zaremba, and
  Pieter Abbeel.
\newblock Domain randomization for transferring deep neural networks from
  simulation to the real world.
\newblock \emph{2017 IEEE/RSJ International Conference on Intelligent Robots
  and Systems (IROS)}, pages 23--30, September 2017.
\newblock \doi{10.1109/iros.2017.8202133}.
\newblock URL \url{http://dx.doi.org/10.1109/IROS.2017.8202133}.

\bibitem[Sadeghi et~al.(2018)Sadeghi, Toshev, Jang, and Levine]{SadeghiTJL18}
Fereshteh Sadeghi, Alexander Toshev, Eric Jang, and Sergey Levine.
\newblock Sim2real viewpoint invariant visual servoing by recurrent control.
\newblock In \emph{2018 {IEEE} Conference on Computer Vision and Pattern
  Recognition, {CVPR} 2018, Salt Lake City, UT, USA, June 18-22, 2018}, pages
  4691--4699, 2018.
\newblock \doi{10.1109/CVPR.2018.00493}.
\newblock URL
  \url{http://openaccess.thecvf.com/content\_cvpr\_2018/html/Sadeghi\_Sim2Real\_Viewpoint\_Invariant\_CVPR\_2018\_paper.html}.

\bibitem[Tan et~al.(2018)Tan, Zhang, Coumans, Iscen, Bai, Hafner, Bohez, and
  Vanhoucke]{sim2real}
Jie Tan, Tingnan Zhang, Erwin Coumans, Atil Iscen, Yunfei Bai, Danijar Hafner,
  Steven Bohez, and Vincent Vanhoucke.
\newblock Sim-to-real: Learning agile locomotion for quadruped robots.
\newblock \emph{Robotics: Science and Systems XIV}, June 2018.
\newblock \doi{10.15607/rss.2018.xiv.010}.
\newblock URL \url{http://dx.doi.org/10.15607/RSS.2018.XIV.010}.

\bibitem[OpenAI et~al.(2019)OpenAI, Akkaya, Andrychowicz, Chociej, Litwin,
  McGrew, Petron, Paino, Plappert, Powell, Ribas, Schneider, Tezak, Tworek,
  Welinder, Weng, Yuan, Zaremba, and Zhang]{openai2019solving}
OpenAI, Ilge Akkaya, Marcin Andrychowicz, Maciek Chociej, Mateusz Litwin, Bob
  McGrew, Arthur Petron, Alex Paino, Matthias Plappert, Glenn Powell, Raphael
  Ribas, Jonas Schneider, Nikolas Tezak, Jerry Tworek, Peter Welinder, Lilian
  Weng, Qiming Yuan, Wojciech Zaremba, and Lei Zhang.
\newblock Solving rubik's cube with a robot hand, 2019.

\bibitem[Rusu et~al.(2016{\natexlab{b}})Rusu, Vecer{\'i}k, Roth{\"o}rl, Heess,
  Pascanu, and Hadsell]{Rusu2016SimtoRealRL}
Andrei~A. Rusu, Matej Vecer{\'i}k, Thomas Roth{\"o}rl, Nicolas Manfred~Otto
  Heess, Razvan Pascanu, and Raia Hadsell.
\newblock Sim-to-real robot learning from pixels with progressive nets.
\newblock In \emph{CoRL}, 2016{\natexlab{b}}.

\bibitem[Peng et~al.(2018)Peng, Andrychowicz, Zaremba, and Abbeel]{peng2018sim}
Xue~Bin Peng, Marcin Andrychowicz, Wojciech Zaremba, and Pieter Abbeel.
\newblock Sim-to-real transfer of robotic control with dynamics randomization.
\newblock In \emph{2018 IEEE international conference on robotics and
  automation (ICRA)}, pages 1--8. IEEE, 2018.

\bibitem[Higuera et~al.(2017)Higuera, Meger, and Dudek]{higuera2017adapting}
Juan Camilo~Gamboa Higuera, David Meger, and Gregory Dudek.
\newblock Adapting learned robotics behaviours through policy adjustment.
\newblock In \emph{2017 IEEE International Conference on Robotics and
  Automation (ICRA)}, pages 5837--5843. IEEE, 2017.

\bibitem[H{\"a}m{\"a}l{\"a}inen et~al.(2019)H{\"a}m{\"a}l{\"a}inen, Arndt,
  Ghadirzadeh, and Kyrki]{hamalainen2019affordance}
Aleksi H{\"a}m{\"a}l{\"a}inen, Karol Arndt, Ali Ghadirzadeh, and Ville Kyrki.
\newblock Affordance learning for end-to-end visuomotor robot control.
\newblock \emph{arXiv preprint arXiv:1903.04053}, 2019.

\bibitem[Riedmiller et~al.(2018)Riedmiller, Hafner, Lampe, Neunert, Degrave,
  de~Wiele, Mnih, Heess, and Springenberg]{Riedmiller2018LearningBP}
Martin~A. Riedmiller, Roland Hafner, Thomas Lampe, Michael Neunert, Jonas
  Degrave, Tom~Van de~Wiele, Volodymyr Mnih, Nicolas Manfred~Otto Heess, and
  Jost~Tobias Springenberg.
\newblock Learning by playing solving sparse reward tasks from scratch.
\newblock In \emph{ICML}, 2018.

\bibitem[Mirowski et~al.(2016)Mirowski, Pascanu, Viola, Soyer, Ballard, Banino,
  Denil, Goroshin, Sifre, Kavukcuoglu, et~al.]{mirowski2016learning}
Piotr Mirowski, Razvan Pascanu, Fabio Viola, Hubert Soyer, Andrew~J Ballard,
  Andrea Banino, Misha Denil, Ross Goroshin, Laurent Sifre, Koray Kavukcuoglu,
  et~al.
\newblock Learning to navigate in complex environments.
\newblock \emph{arXiv preprint arXiv:1611.03673}, 2016.

\bibitem[Sax et~al.(2019)Sax, Emi, Zamir, Guibas, Savarese, and
  Malik]{sax2019mid}
Alexander Sax, Bradley Emi, Amir~R. Zamir, Leonidas~J. Guibas, Silvio Savarese,
  and Jitendra Malik.
\newblock Mid-level visual representations improve generalization and sample
  efficiency for learning visuomotor policies.
\newblock In \emph{Conference on Robot Learning}, 2019.

\bibitem[Sermanet et~al.(2017)Sermanet, Xu, and Levine]{Sermanet2017Rewards}
Pierre Sermanet, Kelvin Xu, and Sergey Levine.
\newblock Unsupervised perceptual rewards for imitation learning.
\newblock \emph{Proceedings of Robotics: Science and Systems (RSS)}, 2017.
\newblock URL \url{http://arxiv.org/abs/1612.06699}.

\bibitem[Hazara and Kyrki(2019)]{hazara2019transferring}
Murtaza Hazara and Ville Kyrki.
\newblock Transferring generalizable motor primitives from simulation to real
  world.
\newblock \emph{IEEE Robotics and Automation Letters}, 4\penalty0 (2):\penalty0
  2172--2179, 2019.

\bibitem[Nair et~al.(2020)Nair, Dalal, Gupta, and Levine]{nair2020accelerating}
Ashvin Nair, Murtaza Dalal, Abhishek Gupta, and Sergey Levine.
\newblock Accelerating online reinforcement learning with offline datasets.
\newblock \emph{arXiv preprint arXiv:2006.09359}, 2020.

\bibitem[Thrun and Mitchell(1995)]{thrun1995lifelong}
Sebastian Thrun and Tom~M Mitchell.
\newblock Lifelong robot learning.
\newblock \emph{Robotics and autonomous systems}, 15\penalty0 (1-2):\penalty0
  25--46, 1995.

\bibitem[Taylor et~al.(2007)Taylor, Stone, and Liu]{taylor2007transfer}
Matthew~E Taylor, Peter Stone, and Yaxin Liu.
\newblock Transfer learning via inter-task mappings for temporal difference
  learning.
\newblock \emph{Journal of Machine Learning Research}, 8\penalty0 (9), 2007.

\bibitem[Cao et~al.(2021)Cao, Kwon, and Sadigh]{cao2021transfer}
Zhangjie Cao, Minae Kwon, and Dorsa Sadigh.
\newblock Transfer reinforcement learning across homotopy classes.
\newblock \emph{IEEE Robotics and Automation Letters}, 6\penalty0 (2):\penalty0
  2706--2713, 2021.

\bibitem[Bodnar et~al.(2020)Bodnar, Hausman, Dulac-Arnold, and
  Jonschkowski]{bodnar2020geometric}
Cristian Bodnar, Karol Hausman, Gabriel Dulac-Arnold, and Rico Jonschkowski.
\newblock A geometric perspective on self-supervised policy adaptation.
\newblock \emph{arXiv preprint arXiv:2011.07318}, 2020.

\bibitem[Kumar et~al.(2020)Kumar, Kumar, Levine, and Finn]{kumar2020one}
Saurabh Kumar, Aviral Kumar, Sergey Levine, and Chelsea Finn.
\newblock One solution is not all you need: Few-shot extrapolation via
  structured maxent rl.
\newblock \emph{Advances in Neural Information Processing Systems}, 33, 2020.

\bibitem[Luna~Gutierrez and Leonetti(2020)]{luna2020information}
R~Luna~Gutierrez and M~Leonetti.
\newblock Information-theoretic task selection for meta-reinforcement learning.
\newblock In \emph{Proceedings of the 34th Conference on Neural Information
  Processing Systems (NeurIPS 2020)}. Leeds, 2020.

\bibitem[Yen-Chen et~al.(2020)Yen-Chen, Zeng, Song, Isola, and
  Lin]{yen2020learning}
Lin Yen-Chen, Andy Zeng, Shuran Song, Phillip Isola, and Tsung-Yi Lin.
\newblock Learning to see before learning to act: Visual pre-training for
  manipulation.
\newblock In \emph{2020 IEEE International Conference on Robotics and
  Automation (ICRA)}, pages 7286--7293. IEEE, 2020.

\bibitem[Khetarpal et~al.(2020)Khetarpal, Riemer, Rish, and
  Precup]{khetarpal2020towards}
Khimya Khetarpal, Matthew Riemer, Irina Rish, and Doina Precup.
\newblock Towards continual reinforcement learning: A review and perspectives.
\newblock \emph{arXiv preprint arXiv:2012.13490}, 2020.

\bibitem[Pastor et~al.(2012)Pastor, Kalakrishnan, Righetti, and
  Schaal]{pastor2012asm}
Peter Pastor, Mrinal Kalakrishnan, Ludovic Righetti, and Stefan Schaal.
\newblock Towards associative skill memories.
\newblock In \emph{2012 12th IEEE-RAS International Conference on Humanoid
  Robots (Humanoids 2012)}, pages 309--315, November 2012.
\newblock \doi{10.1109/HUMANOIDS.2012.6651537}.

\bibitem[Rueckert et~al.(2015)Rueckert, Mundo, Paraschos, Peters, and
  Neumann]{rueckert2015movprim}
E.~Rueckert, J.~Mundo, A.~Paraschos, J.~Peters, and G.~Neumann.
\newblock Extracting low-dimensional control variables for movement primitives.
\newblock In \emph{2015 IEEE International Conference on Robotics and
  Automation (ICRA)}, pages 1511--1518, May 2015.
\newblock \doi{10.1109/ICRA.2015.7139390}.

\bibitem[Zhou et~al.(2020)Zhou, Wu, Rojas, Xu, and Li]{zhou2020incremental}
Xuefeng Zhou, Hongmin Wu, Juan Rojas, Zhihao Xu, and Shuai Li.
\newblock Incremental learning robot task representation and identification.
\newblock In \emph{Nonparametric Bayesian Learning for Collaborative Robot
  Multimodal Introspection}, pages 29--49. Springer, 2020.

\bibitem[Tanneberg et~al.(2021)Tanneberg, Ploeger, Rueckert, and
  Peters]{tanneberg2021skid}
Daniel Tanneberg, Kai Ploeger, Elmar Rueckert, and Jan Peters.
\newblock Skid raw: Skill discovery from raw trajectories.
\newblock \emph{IEEE Robotics and Automation Letters}, 2021.

\bibitem[Yang et~al.(2020)Yang, Xu, Wu, and Wang]{yang2020multi}
Ruihan Yang, Huazhe Xu, Yi~Wu, and Xiaolong Wang.
\newblock Multi-task reinforcement learning with soft modularization.
\newblock \emph{arXiv preprint arXiv:2003.13661}, 2020.

\bibitem[Ichter et~al.(2020)Ichter, Sermanet, and Lynch]{ichter2020broadly}
Brian Ichter, Pierre Sermanet, and Corey Lynch.
\newblock Broadly-exploring, local-policy trees for long-horizon task planning.
\newblock \emph{arXiv preprint arXiv:2010.06491}, 2020.

\bibitem[Wulfmeier et~al.(2020)Wulfmeier, Rao, Hafner, Lampe, Abdolmaleki,
  Hertweck, Neunert, Tirumala, Siegel, Heess, et~al.]{wulfmeier2020data}
Markus Wulfmeier, Dushyant Rao, Roland Hafner, Thomas Lampe, Abbas Abdolmaleki,
  Tim Hertweck, Michael Neunert, Dhruva Tirumala, Noah Siegel, Nicolas Heess,
  et~al.
\newblock Data-efficient hindsight off-policy option learning.
\newblock \emph{arXiv preprint arXiv:2007.15588}, 2020.

\bibitem[Vezzani et~al.(2020)Vezzani, Neunert, Wulfmeier, Jeong, Lampe, Siegel,
  Hafner, Abdolmaleki, Riedmiller, and Nori]{vezzani2020not}
Giulia Vezzani, Michael Neunert, Markus Wulfmeier, Rae Jeong, Thomas Lampe,
  Noah Siegel, Roland Hafner, Abbas Abdolmaleki, Martin Riedmiller, and
  Francesco Nori.
\newblock " what, not how"--solving an under-actuated insertion task from
  scratch.
\newblock \emph{arXiv preprint arXiv:2010.15492}, 2020.

\bibitem[Camacho et~al.(2020)Camacho, Varley, Jain, Iscen, and
  Kalashnikov]{camacho2020disentangled}
Alberto Camacho, Jacob Varley, Deepali Jain, Atil Iscen, and Dmitry
  Kalashnikov.
\newblock Disentangled planning and control in vision based robotics via reward
  machines.
\newblock \emph{arXiv preprint arXiv:2012.14464}, 2020.

\bibitem[Li et~al.(2021)Li, Wu, Xu, Wang, and Wu]{li2021solving}
Yunfei Li, Yilin Wu, Huazhe Xu, Xiaolong Wang, and Yi~Wu.
\newblock Solving compositional reinforcement learning problems via task
  reduction.
\newblock \emph{arXiv preprint arXiv:2103.07607}, 2021.

\bibitem[Lu et~al.(2021)Lu, Shen, Zhou, Courville, Tenenbaum, and
  Gan]{lu2021learning}
Yuchen Lu, Yikang Shen, Siyuan Zhou, Aaron Courville, Joshua~B Tenenbaum, and
  Chuang Gan.
\newblock Learning task decomposition with ordered memory policy network.
\newblock \emph{arXiv preprint arXiv:2103.10972}, 2021.

\bibitem[Kroemer et~al.(2015)Kroemer, Daniel, Neumann, Van~Hoof, and
  Peters]{kroemer2015towards}
Oliver Kroemer, Christian Daniel, Gerhard Neumann, Herke Van~Hoof, and Jan
  Peters.
\newblock Towards learning hierarchical skills for multi-phase manipulation
  tasks.
\newblock In \emph{2015 IEEE International Conference on Robotics and
  Automation (ICRA)}, pages 1503--1510. IEEE, 2015.

\bibitem[Peng et~al.(2019)Peng, Chang, Zhang, Abbeel, and Levine]{peng2019mcp}
Xue~Bin Peng, Michael Chang, Grace Zhang, Pieter Abbeel, and Sergey Levine.
\newblock Mcp: Learning composable hierarchical control with multiplicative
  compositional policies.
\newblock \emph{arXiv preprint arXiv:1905.09808}, 2019.

\bibitem[Hasenclever et~al.(2020)Hasenclever, Pardo, Hadsell, Heess, and
  Merel]{hasenclever2020complementary}
Leonard Hasenclever, Fabio Pardo, Raia Hadsell, Nicolas Heess, and Josh Merel.
\newblock {C}o{M}ic: Complementary task learning \& mimicry for reusable
  skills.
\newblock In Hal~Daumé III and Aarti Singh, editors, \emph{Proceedings of the
  37th International Conference on Machine Learning}, volume 119 of
  \emph{Proceedings of Machine Learning Research}, pages 4105--4115. PMLR,
  13--18 Jul 2020.
\newblock URL \url{http://proceedings.mlr.press/v119/hasenclever20a.html}.

\bibitem[Merel et~al.(2020)Merel, Tunyasuvunakool, Ahuja, Tassa, Hasenclever,
  Pham, Erez, Wayne, and Heess]{merel2020catch}
Josh Merel, Saran Tunyasuvunakool, Arun Ahuja, Yuval Tassa, Leonard
  Hasenclever, Vu~Pham, Tom Erez, Greg Wayne, and Nicolas Heess.
\newblock Catch \& carry: reusable neural controllers for vision-guided
  whole-body tasks.
\newblock \emph{ACM Transactions on Graphics (TOG)}, 39\penalty0 (4):\penalty0
  39--1, 2020.

\bibitem[Li et~al.(2020)Li, Lambert, Calandra, Meier, and Rai]{li2020learning}
Tianyu Li, Nathan Lambert, Roberto Calandra, Franziska Meier, and Akshara Rai.
\newblock Learning generalizable locomotion skills with hierarchical
  reinforcement learning.
\newblock In \emph{2020 IEEE International Conference on Robotics and
  Automation (ICRA)}, pages 413--419. IEEE, 2020.

\bibitem[Tirumala et~al.(2020)Tirumala, Galashov, Noh, Hasenclever, Pascanu,
  Schwarz, Desjardins, Czarnecki, Ahuja, Teh, et~al.]{tirumala2020behavior}
Dhruva Tirumala, Alexandre Galashov, Hyeonwoo Noh, Leonard Hasenclever, Razvan
  Pascanu, Jonathan Schwarz, Guillaume Desjardins, Wojciech~Marian Czarnecki,
  Arun Ahuja, Yee~Whye Teh, et~al.
\newblock Behavior priors for efficient reinforcement learning.
\newblock \emph{arXiv preprint arXiv:2010.14274}, 2020.

\bibitem[Hausman et~al.(2018)Hausman, Springenberg, Wang, Heess, and
  Riedmiller]{hausman2018learning}
Karol Hausman, Jost~Tobias Springenberg, Ziyu Wang, Nicolas Heess, and Martin
  Riedmiller.
\newblock Learning an embedding space for transferable robot skills.
\newblock In \emph{International Conference on Learning Representations}, 2018.
\newblock URL \url{https://openreview.net/forum?id=rk07ZXZRb}.

\bibitem[Julian et~al.(2018)Julian, Heiden, He, Zhang, Schaal, Lim, Sukhatme,
  and Hausman]{julian2018scaling}
Ryan~C Julian, Eric Heiden, Zhanpeng He, Hejia Zhang, Stefan Schaal, Joseph
  Lim, Gaurav~S Sukhatme, and Karol Hausman.
\newblock Scaling simulation-to-real transfer by learning composable robot
  skills.
\newblock In \emph{International Symposium on Experimental Robotics}. Springer,
  2018.
\newblock URL \url{https://ryanjulian.me/iser\_2018.pdf}.

\bibitem[Benureau and Oudeyer(2016)]{benureau2016behavioral}
Fabien~CY Benureau and Pierre-Yves Oudeyer.
\newblock Behavioral diversity generation in autonomous exploration through
  reuse of past experience.
\newblock \emph{Frontiers in Robotics and AI}, 3:\penalty0 8, 2016.

\bibitem[Singh et~al.(2020{\natexlab{a}})Singh, Liu, Zhou, Yu, Rhinehart, and
  Levine]{singh2020parrot}
Avi Singh, Huihan Liu, Gaoyue Zhou, Albert Yu, Nicholas Rhinehart, and Sergey
  Levine.
\newblock Parrot: Data-driven behavioral priors for reinforcement learning.
\newblock \emph{arXiv preprint arXiv:2011.10024}, 2020{\natexlab{a}}.

\bibitem[Biza et~al.(2021)Biza, Wang, Platt, van~de Meent, and
  Wong]{biza2021action}
Ondrej Biza, Dian Wang, Robert Platt, Jan-Willem van~de Meent, and Lawson~LS
  Wong.
\newblock Action priors for large action spaces in robotics.
\newblock \emph{arXiv preprint arXiv:2101.04178}, 2021.

\bibitem[Singh et~al.(2020{\natexlab{b}})Singh, Yu, Yang, Zhang, Kumar, and
  Levine]{singh2020cog}
Avi Singh, Albert Yu, Jonathan Yang, Jesse Zhang, Aviral Kumar, and Sergey
  Levine.
\newblock Cog: Connecting new skills to past experience with offline
  reinforcement learning.
\newblock \emph{arXiv preprint arXiv:2010.14500}, 2020{\natexlab{b}}.

\bibitem[Allshire et~al.(2021)Allshire, Mart{\'\i}n-Mart{\'\i}n, Lin, Manuel,
  Savarese, and Garg]{allshire2021laser}
Arthur Allshire, Roberto Mart{\'\i}n-Mart{\'\i}n, Charles Lin, Shawn Manuel,
  Silvio Savarese, and Animesh Garg.
\newblock Laser: Learning a latent action space for efficient reinforcement
  learning.
\newblock \emph{arXiv preprint arXiv:2103.15793}, 2021.

\bibitem[Mendez et~al.(2020)Mendez, Wang, and Eaton]{mendez2020lifelong}
Jorge Mendez, Boyu Wang, and Eric Eaton.
\newblock Lifelong policy gradient learning of factored policies for faster
  training without forgetting.
\newblock \emph{Advances in Neural Information Processing Systems}, 33, 2020.

\bibitem[Lu et~al.(2020)Lu, Grover, Abbeel, and Mordatch]{lu2020reset}
Kevin Lu, Aditya Grover, Pieter Abbeel, and Igor Mordatch.
\newblock Reset-free lifelong learning with skill-space planning.
\newblock \emph{arXiv preprint arXiv:2012.03548}, 2020.

\bibitem[Koenig and Matari{\'c}(2017)]{koenig2017robot}
Nathan Koenig and Maja~J Matari{\'c}.
\newblock Robot life-long task learning from human demonstrations: a bayesian
  approach.
\newblock \emph{Autonomous Robots}, 41\penalty0 (5):\penalty0 1173--1188, 2017.

\bibitem[Hazara et~al.(2019)Hazara, Li, and Kyrki]{hazara2019active}
Murtaza Hazara, Xiaopu Li, and Ville Kyrki.
\newblock Active incremental learning of a contextual skill model.
\newblock In \emph{2019 IEEE/RSJ International Conference on Intelligent Robots
  and Systems (IROS)}, pages 1834--1839. IEEE, 2019.

\bibitem[Hawasly and Ramamoorthy(2013)]{hawasly2013lifelong}
Majd Hawasly and Subramanian Ramamoorthy.
\newblock Lifelong transfer learning with an option hierarchy.
\newblock In \emph{2013 IEEE/RSJ International Conference on Intelligent Robots
  and Systems}, pages 1341--1346. IEEE, 2013.

\bibitem[Xiong et~al.(2021)Xiong, Liu, Huang, Yang, and Qiao]{xiong2021state}
Fangzhou Xiong, Zhiyong Liu, Kaizhu Huang, Xu~Yang, and Hong Qiao.
\newblock State primitive learning to overcome catastrophic forgetting in
  robotics.
\newblock \emph{Cognitive Computation}, 13\penalty0 (2):\penalty0 394--402,
  2021.

\bibitem[Maeda et~al.(2017)Maeda, Ewerton, Osa, Busch, and
  Peters]{maeda2017active}
Guilherme Maeda, Marco Ewerton, Takayuki Osa, Baptiste Busch, and Jan Peters.
\newblock Active incremental learning of robot movement primitives.
\newblock In Sergey Levine, Vincent Vanhoucke, and Ken Goldberg, editors,
  \emph{Proceedings of the 1st Annual Conference on Robot Learning}, volume~78
  of \emph{Proceedings of Machine Learning Research}, pages 37--46. PMLR,
  13--15 Nov 2017.
\newblock URL \url{http://proceedings.mlr.press/v78/maeda17a.html}.

\bibitem[Traor{\'e} et~al.(2019)Traor{\'e}, Caselles-Dupr{\'e}, Lesort, Sun,
  Cai, D{\'\i}az-Rodr{\'\i}guez, and Filliat]{traore2019discorl}
Ren{\'e} Traor{\'e}, Hugo Caselles-Dupr{\'e}, Timoth{\'e}e Lesort, Te~Sun,
  Guanghang Cai, Natalia D{\'\i}az-Rodr{\'\i}guez, and David Filliat.
\newblock Discorl: Continual reinforcement learning via policy distillation.
\newblock \emph{arXiv preprint arXiv:1907.05855}, 2019.

\bibitem[Stulp(2012)]{stulp2012adaptive}
Freek Stulp.
\newblock Adaptive exploration for continual reinforcement learning.
\newblock In \emph{2012 IEEE/RSJ International Conference on Intelligent Robots
  and Systems}, pages 1631--1636. IEEE, 2012.

\bibitem[Fern{\'a}ndez and Veloso(2006)]{fernandez2006probabilistic}
Fernando Fern{\'a}ndez and Manuela Veloso.
\newblock Probabilistic policy reuse in a reinforcement learning agent.
\newblock In \emph{Proceedings of the fifth international joint conference on
  Autonomous agents and multiagent systems}, pages 720--727, 2006.

\bibitem[Sharma et~al.(2020)Sharma, Liang, Zhao, LaGrassa, and
  Kroemer]{sharma2020learning}
Mohit Sharma, Jacky Liang, Jialiang Zhao, Alex LaGrassa, and Oliver Kroemer.
\newblock Learning to compose hierarchical object-centric controllers for
  robotic manipulation.
\newblock \emph{arXiv preprint arXiv:2011.04627}, 2020.

\bibitem[Raziei and Moghaddam(2020)]{raziei2020adaptable}
Zohreh Raziei and Mohsen Moghaddam.
\newblock Adaptable automation with modular deep reinforcement learning and
  policy transfer.
\newblock \emph{arXiv preprint arXiv:2012.01934}, 2020.

\bibitem[Narvekar et~al.(2020)Narvekar, Peng, Leonetti, Sinapov, Taylor, and
  Stone]{narvekar2020curriculum}
Sanmit Narvekar, Bei Peng, Matteo Leonetti, Jivko Sinapov, Matthew~E Taylor,
  and Peter Stone.
\newblock Curriculum learning for reinforcement learning domains: A framework
  and survey.
\newblock \emph{Journal of Machine Learning Research}, 21\penalty0
  (181):\penalty0 1--50, 2020.

\bibitem[Fabisch et~al.(2015)Fabisch, Metzen, Krell, and
  Kirchner]{fabisch2015accounting}
Alexander Fabisch, Jan~Hendrik Metzen, Mario~Michael Krell, and Frank Kirchner.
\newblock Accounting for task-difficulty in active multi-task robot control
  learning.
\newblock \emph{KI-K{\"u}nstliche Intelligenz}, 29\penalty0 (4):\penalty0
  369--377, 2015.

\bibitem[Foglino et~al.(2019)Foglino, Coletto~Christakou, Luna~Gutierrez, and
  Leonetti]{foglino2019curriculum}
F~Foglino, C~Coletto~Christakou, R~Luna~Gutierrez, and M~Leonetti.
\newblock Curriculum learning for cumulative return maximization.
\newblock In \emph{Proceedings of the 28th International Joint Conference on
  Artificial Intelligence}, pages 2308--2314. IJCAI, 2019.

\bibitem[Sinapov et~al.(2015)Sinapov, Narvekar, Leonetti, and
  Stone]{sinapov2015learning}
Jivko Sinapov, Sanmit Narvekar, Matteo Leonetti, and Peter Stone.
\newblock Learning inter-task transferability in the absence of target task
  samples.
\newblock In \emph{Proceedings of the 2015 International Conference on
  Autonomous Agents and Multiagent Systems}, pages 725--733, 2015.

\bibitem[Kruskal(1956)]{kruskal1956shortest}
Joseph~B Kruskal.
\newblock On the shortest spanning subtree of a graph and the traveling
  salesman problem.
\newblock \emph{Proceedings of the American Mathematical society}, 7\penalty0
  (1):\penalty0 48--50, 1956.

\end{thebibliography}
